# Topic Segmentation and Labeling
# in Asynchronous Conversations

**Shafiq Joty**                                                        SJOTY@QF.ORG.QA
*Qatar Computing Research Institute*
*Qatar Foundation*
*Doha, Qatar*

**Giuseppe Carenini**                                               CARENINI@CS.UBC.CA
**Raymond T. Ng**                                                        RNG@CS.UBC.CA
*University of British Columbia*
*Vancouver, BC, Canada, V6T 1Z4*

## Abstract

Topic segmentation and labeling is often considered a prerequisite for higher-level conversation analysis and has been shown to be useful in many Natural Language Processing (NLP) applications. We present two new corpora of email and blog conversations annotated with topics, and evaluate annotator reliability for the segmentation and labeling tasks in these asynchronous conversations. We propose a complete computational framework for topic segmentation and labeling in asynchronous conversations. Our approach extends state-of-the-art methods by considering a fine-grained structure of an asynchronous conversation, along with other conversational features by applying recent graph-based methods for NLP. For topic segmentation, we propose two novel unsupervised models that exploit the fine-grained conversational structure, and a novel graph-theoretic supervised model that combines lexical, conversational and topic features. For topic labeling, we propose two novel (unsupervised) random walk models that respectively capture conversation specific clues from two different sources: the leading sentences and the fine-grained conversational structure. Empirical evaluation shows that the segmentation and the labeling performed by our best models beat the state-of-the-art, and are highly correlated with human annotations.

## 1. Introduction

With the ever increasing popularity of Internet technologies, it is very common nowadays for people to discuss events, issues, tasks and personal experiences in social media (e.g., Facebook, Twitter, blogs, fora) and email (Verna, 2010; Baron, 2008). These are examples of **asynchronous conversations** where participants communicate with each other at different times. The huge amount of textual data generated everyday in these conversations calls for automated methods of conversational text analysis. Effective processing of these conversational texts can be of great strategic value for both organizations and individuals (Carenini, Murray, & Ng, 2011). For instance, managers can find the information exchanged in email conversations within a company to be extremely valuable for decision auditing. If a decision turns out to be ill-advised, mining the relevant conversations may help in determining responsibility and accountability. Similarly, conversations that led to favorable decisions could be mined to identify effective communication patterns and sources within the company. In public blogging services (e.g., Twitter, Slashdot), conversations of-ten get very large involving hundreds of bloggers making potentially thousands of comments. During a major event such as a political uprising in Egypt, relevant messages are posted by the thousands or millions. It is simply not feasible to read all messages relevant to such an event, and so mining and summarization technologies can help providing an overview of what people are saying and what positive or negative opinions are being expressed. Mining and summarizing of conversations also aid improved indexing and searching. On a more personal level, an informative summary of a conversation could greatly support a new participant to get up to speed and join an already existing conversation. It could also help someone to quickly prepare for a follow-up discussion of a conversation she was already part of, but which occurred too long ago for her to remember the details.

**Topic segmentation and labeling** is often considered a prerequisite for higher-level conversation analysis (Bangalore, Di Fabbrizio, & Stent, 2006) and has been shown to be useful in many Natural Language Processing (NLP) applications including automatic summarization (Harabagiu & Lacatusu, 2005; Kleinbauer, Becker, & Becker, 2007; Dias, Alves, & Lopes, 2007), text generation (Barzilay & Lee, 2004), information extraction (Allan, 2002), and conversation visualization (Liu, Zhou, Pan, Song, Qian, Cai, & Lian, 2012).

Adapting the standard definition of topic (Galley, McKeown, Fosler-Lussier, & Jing, 2003) to asynchronous conversations, we consider a topic to be something about which the participants discuss or argue or express their opinions. Multiple topics seem to occur naturally in social interactions, whether synchronous (e.g., meetings, chats) or asynchronous (e.g., emails, blogs). In the naturally occurring ICSI multi-party meetings (Janin et al. 2003), Galley et al. (2003) report an average of 7.5 topical segments per conversation. In multi-party chat, Elsner and Charniak (2010) report an average of 2.75 discussions active at a time. In the email and blog corpora, that we present in this article, annotators found an average of 2.5 and 10.77 topics per email and blog conversation, respectively.

**Topic segmentation** refers to the task of grouping the sentences of an asynchronous conversation into a set of coherent topical clusters (or segments)[1], and **topic labeling** is the task of assigning a short description to each of the topical clusters to facilitate interpretations of the topics (Purver, 2011). For example, in the sample truncated email conversation from our corpora shown in Figure 1, the majority of our three annotators found three different topics (or clusters). Likewise, in the truncated blog conversation shown in Figure 2, our annotators found six different topics. The right column in each figure specifies a particular segmentation by assigning the same topic ID (or cluster ID) to sentences belonging to the same topic. The topics in each figure are also differentiated using different colors. The topic labels assigned by the annotators are listed below each conversation (e.g., 'Telecon cancellation', 'Tag document', 'Responding to I18N' in Figure 1).

While extensive research has been conducted in topic segmentation for monolog (e.g., news articles) and synchronous dialog (e.g., meetings), none has studied the problem of segmenting and labeling asynchronous conversations (e.g., email, blog). Therefore, there is no reliable annotation scheme, no standard corpus, and no agreed-upon metrics available. Also, it is our key observation that, because of its asynchronous nature, and the use of quotation (Crystal, 2001), topics in these conversations are often interleaved and do not change in a sequential way. That is, if we look at the temporal order of the sentences in

---

1. In this article, we use the terms topical cluster and topical segment interchangeably.

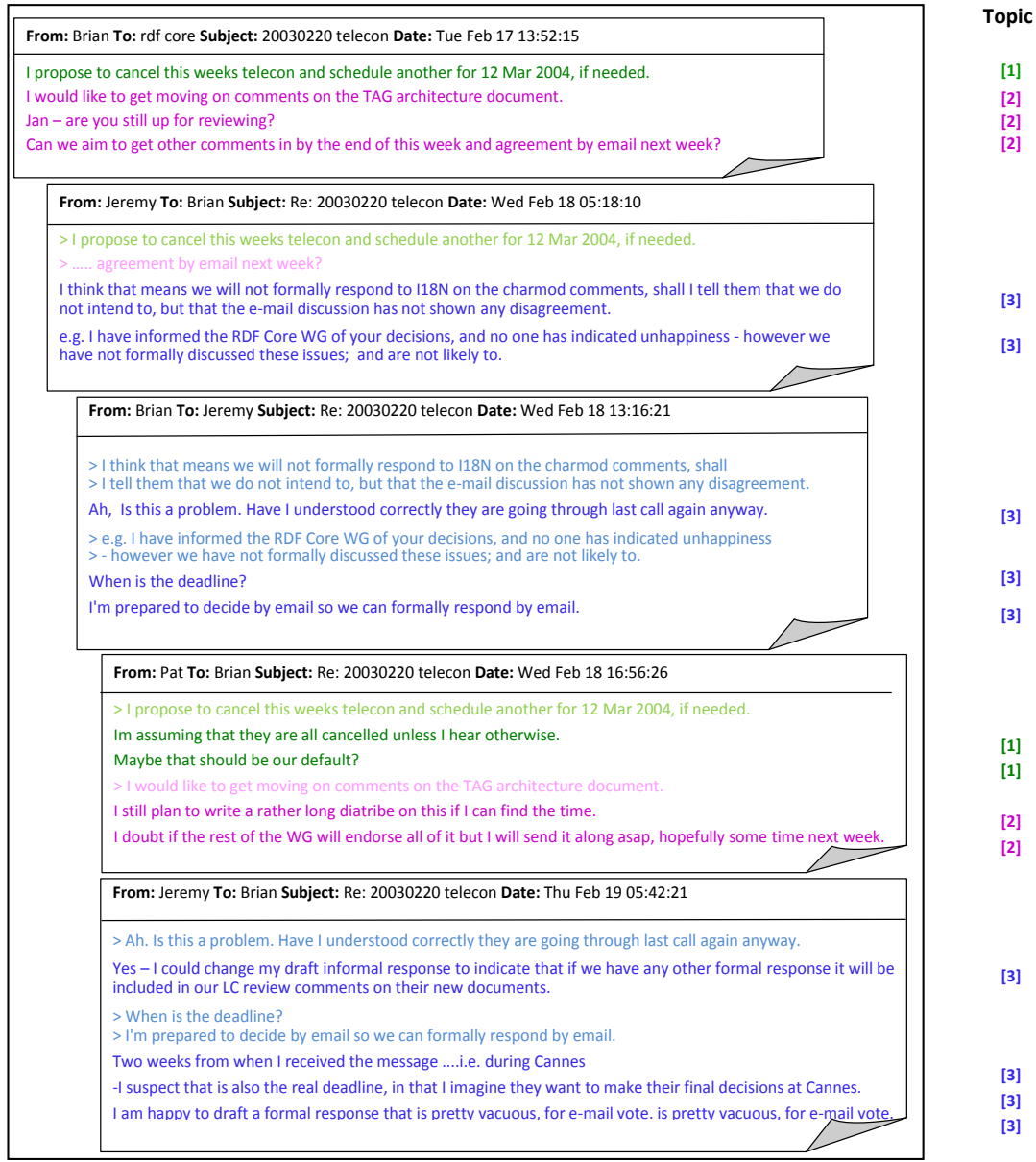

**Topic**

**From:** Brian **To:** rdf core **Subject:** 20030220 telecon **Date:** Tue Feb 17 13:52:15

I propose to cancel this weeks telecon and schedule another for 12 Mar 2004, if needed. **[1]**
I would like to get moving on comments on the TAG architecture document. **[2]**
Jan – are you still up for reviewing? **[2]**
Can we aim to get other comments in by the end of this week and agreement by email next week? **[2]**

**From:** Jeremy **To:** Brian **Subject:** Re: 20030220 telecon **Date:** Wed Feb 18 05:18:10

> I propose to cancel this weeks telecon and schedule another for 12 Mar 2004, if needed.
> ..... agreement by email next week?
I think that means we will not formally respond to I18N on the charmod comments, shall I tell them that we do **[3]**
not intend to, but that the e-mail discussion has not shown any disagreement.
e.g. I have informed the RDF Core WG of your decisions, and no one has indicated unhappiness - however we **[3]**
have not formally discussed these issues; and are not likely to.

**From:** Brian **To:** Jeremy **Subject:** Re: 20030220 telecon **Date:** Wed Feb 18 13:16:21

> I think that means we will not formally respond to I18N on the charmod comments, shall
> I tell them that we do not intend to, but that the e-mail discussion has not shown any disagreement.
Ah,  Is this a problem. Have I understood correctly they are going through last call again anyway. **[3]**
> e.g. I have informed the RDF Core WG of your decisions, and no one has indicated unhappiness
> - however we have not formally discussed these issues; and are not likely to.
When is the deadline? **[3]**
I'm prepared to decide by email so we can formally respond by email. **[3]**

**From:** Pat **To:** Brian **Subject:** Re: 20030220 telecon **Date:** Wed Feb 18 16:56:26

> I propose to cancel this weeks telecon and schedule another for 12 Mar 2004, if needed.
Im assuming that they are all cancelled unless I hear otherwise. **[1]**
Maybe that should be our default? **[1]**
> I would like to get moving on comments on the TAG architecture document.
I still plan to write a rather long diatribe on this if I can find the time. **[2]**
I doubt if the rest of the WG will endorse all of it but I will send it along asap, hopefully some time next week. **[2]**

**From:** Jeremy **To:** Brian **Subject:** Re: 20030220 telecon **Date:** Thu Feb 19 05:42:21

> Ah. Is this a problem. Have I understood correctly they are going through last call again anyway.
Yes – I could change my draft informal response to indicate that if we have any other formal response it will be **[3]**
included in our LC review comments on their new documents.
> When is the deadline?
> I'm prepared to decide by email so we can formally respond by email.
Two weeks from when I received the message ....i.e. during Cannes **[3]**
-I suspect that is also the real deadline, in that I imagine they want to make their final decisions at Cannes. **[3]**
I am happy to draft a formal response that is pretty vacuous, for e-mail vote, is pretty vacuous, for e-mail vote. **[3]**

**Topic Labels**

**Topic 1 (green): Telecon cancellation, Topic 2 (magenta): TAG document, Topic 3 (blue): Responding to I18N.**

Figure 1: Sample truncated email conversation from our email corpus. Each color indicates a different topic. The right most column specifies the topic assignments for the sentences.

**Fragment** **Topic**

**Author:** Soulskill **Title:** Bethesda Releases Daggerfall For Free **Type:** Article

On Thursday, Bethesda announced that for the 15th anniversary of the Elder Scrolls series, they were releasing The Elder Scrolls II: Daggerfall for free. **(a)** **[1]**

They aren't providing support for the game anymore, but they posted a detailed description of how to get the game running in DOSBox. **[1]**

Fans of the series can now easily relive the experience of getting completely lost in those enormous dungeons. **(b)** **[2]**
Save often. **[2]**

**Author:** Datamonstar **Title:** Nice nice nice nice... **Comment id:** 1 **Parent id:** None **Type:** Comment

>Fans of the series can now easily relive the experience of getting completely lost in those enormous dungeons.
>Save often.
... well not really, since this game is soooo old, but still its a huge HUGE gameworld. **[2]**
Really, It's big. **(c)** **[2]**
Can't wait to play it. **[2]**
It makes Oblivion look like Sesame Street. **[2]**

**Author:** Freetardo **Title:** Re: Nice nice nice nice... **Comment id:** 2 **Parent id:** 1 **Type:** Comment

Yes it is big, but most of it is just the same thing over and over again. **(d)** **[3]**
It was quite monotonous at times, really. **[3]**

**Author:** gbarules2999 **Title:** Re: Nice nice nice nice... **Comment id:** 3 **Parent id:** 1 **Type:** Comment

Randomly generated HUGE isn't nearly as good as designed small. **(e)** **[4]**
Back to Morrowind, folks. **(f)** **[5]**

**Author:** drinkypoo **Title:** Re: Nice nice nice nice... **Comment id:** 4 **Parent id:** 3 **Type:** Comment

>Randomly generated HUGE isn't nearly as good as designed small.
The solution is obviously to combine both approaches. **[4]**
That way a single game will satisfy both types of players. **(g)** **[4]**

**Author:** ElrondHubbard **Title:** Rest well this night -- **Comment id:** 5 **Parent id:** None **Type:** Comment

-- for tomorrow, you sail for the kingdom... of Daggerfall. **(h)** **[1]**
Many, many enjoyable hours I spent playing this game when I could (should) have been working on my thesis. **(i)** **[1]**
Chief complaint: The repetitive dungeons, stitched together seemingly near-randomly from prefabbed bits and pieces that were repeated endlessly. **(j)** **[3]**
**[3]**
Still, a great game. **[1]**

**Author:** Anonymous **Title:** Re:Rest well this night -- **Comment id:** 6 **Parent id:** 5 **Type:** Comment

>Many, many enjoyable hours I spent playing this game when I could (should) have been working on my thesis
So, how did your thesis go? **(k)** **[0]**
>Chief complaint: The repetitive dungeons, stitched together seemingly near-randomly ...a great game
I also think this is a great game. **(l)** **[1]**

**Topic Labels**

Topic 1 (green): Free release of Daggerfall and reaction, Topic 2 (purple): Game contents or size, Topic 3 (orange): Bugs or faults, Topic 4 (magenta): Game design, Topic 5 (blue): Other gaming options, Topic 0 (red): `OFF-TOPIC'.

Figure 2: Sample truncated blog conversation from our blog corpus. Each color indicates a different topic. The right most column (i.e., Topic) specifies the topic assignments for the sentences. The Fragment column specifies the fragments in the FQG (see Section 3.1.3).

a conversation, the discussion of a topic may appear to intersect with the discussion of others. As can be seen in Figure 1, after a discussion of topic 3 in the second and third email, topics 1 and 2 are revisited in the fourth email, then topic 3 is again brought back in the fifth email. Therefore, the sequentiality constraint of topic segmentation in monolog and synchronous dialog does not hold in asynchronous conversation. As a result, we do not expect models which have proved successful in monolog or synchronous dialog to be as effective, when directly applied to asynchronous conversation.

Our contributions in this article aim to remedy these problems. First, we present two new corpora of email and blog conversations annotated with topics, and evaluate annotator reliability for the topic segmentation and labeling tasks using a new set of metrics, which are also used to evaluate the computational models. To our knowledge, these are the first such corpora that will be made publicly available. Second, we present a complete topic segmentation and labeling framework for asynchronous conversations. Our approach extends state-of-the-art methods (for monologs and synchronous dialogs) by considering a fine-grained structure of the asynchronous conversation along with other conversational features. In doing so, we apply recent graph-based methods for NLP (Mihalcea & Radev, 2011) such as min-cut and random walk on paragraph, sentence or word graphs.

For topic segmentation, we propose two novel unsupervised models that exploit, in a principled way, the fine-grained conversational structure beyond the lexical information. We also propose a novel graph-theoretic supervised topic segmentation model that combines lexical, conversational, and topic features. For topic labeling, we propose to generate labels using an unsupervised extractive approach that identifies the most representative phrases in the text. Specifically, we propose two novel random walk models that respectively captures two forms of conversation specific information: (i) the fact that the leading sentences in a topical cluster often carry the most informative clues, and (ii) the fine-grained conversational structure. To the best of our knowledge, this is also the first comprehensive study to address the problem of topic segmentation and labeling in asynchronous conversation.

Our framework was tested in a series of experiments. Experimental results in the topic segmentation task show that the unsupervised segmentation models benefit when they consider the finer conversational structure of asynchronous conversations. A comparison of the supervised segmentation model with the unsupervised models reveals that the supervised method, by optimizing the relative weights of the features, outperforms the unsupervised ones even using only a few labeled conversations. Remarkably, the segmentation decisions of the best unsupervised and the supervised models are also highly correlated with human annotations. As for the experiments on the topic labeling task, they show that the random walk model performs better when it exploits the conversation specific clues from the leading sentences and the conversational structure. The evaluation of the end-to-end system also shows promising results in both corpora, when compared with human annotations.

In the rest of this article, after discussing related work in Section 2, we present our segmentation and labeling models in Section 3. We then describe our corpora and evaluation metrics in Section 4. The experiments and analysis are presented in Section 5. We summarize our contributions and consider directions for future work in Section 6.

## 2. Related Work

Three research areas are directly related to our study: topic segmentation, topic labeling, and extracting the conversation structure of asynchronous conversations.

### 2.1 Topic Segmentation

Topic segmentation has been extensively studied both for monologs and synchronous dialogs where the task is to divide the discourse into topically coherent **sequential** segments (for a detailed overview see Purver, 2011). The unsupervised models rely on the discourse cohesion phenomenon, where the intuition is that sentences in a segment are lexically similar to each other but not to sentences in the preceding or the following segment. These approaches mainly differ in how they measure the lexical similarity between sentences.

One such early approach is TextTiling (Hearst, 1997), which still forms the baseline for many recent advancements. It operates in three steps: tokenization, lexical score determination, and depth score computation. In the tokenization step, it forms the fixed length pseudo-sentences, each containing $n$ stemmed words. Then it considers blocks of $k$ pseudo-sentences, and for each gap between two consecutive pseudo-sentences it measures the cosine-based lexical similarity between the adjacent blocks by representing them as vectors of term frequencies. Finally, it measures the depth of the similarity valley for each gap, and assigns the topic boundaries at the appropriate sentence gaps based on a threshold.

When similarity is computed only on the basis of raw term frequency (TF) vectors, it can cause problems because of sparseness, and because it treats the terms independently. Choi, Hastings, and Moore (2001) use Latent Semantic Analysis (LSA) to measure the similarity and show that LSA-based similarity performs better than the raw TF-based similarity. Unlike TextTiling, which uses a threshold to decide on topic boundaries, Choi et al. use divisive clustering to find the topical segments. We use similarity measures based on both TF and LSA as features in our supervised segmentation model.

Another variation of the cohesion-based approach is LCSeg (Galley et al., 2003), which uses lexical chains (Morris & Hirst, 1991). LCSeg first finds the chains based on term repetitions, and weights those based on term frequency and chain length. The cosine similarity between two adjacent blocks' lexical chain vectors is then used as a measure of lexical cohesion in a TextTiling-like algorithm to find the segments. LCSeg achieves results comparable to the previous approaches (e.g., Choi et al., 2001) in both monolog (i.e., newspaper) and synchronous dialog (i.e., meeting). Galley et al. also propose a supervised model for segmenting meeting transcripts. They use a C4.5 probabilistic classifier with lexical and conversational features and show that it outperforms the unsupervised method (LCSeg).

Hsueh, Moore, and Renals (2006) apply the models of Galley et al. (2003) to both the manual transcripts and the ASR (automatic speech recognizer) output of meetings. They perform segmentation at both coarse (topic) and fine (subtopic) levels. At the topic level, they get similar results as Galley et al. – the supervised model outperforming LCSeg. However, at the subtopic level, LCSeg surprisingly outperforms the supervised model indicating that finer topic shifts are better characterized by lexical similarity alone.

In our work, we initially show how LCSeg performs poorly, when applied to the temporal ordering of the asynchronous conversation. This is because, as we mentioned earlier, topics in asynchronous conversations are interleaved and do not change sequentially following the temporal order of the sentences. To address this, we propose a novel extension of LCSeg that leverages the fine conversational structure of asynchronous conversations. We also propose a novel supervised segmentation model for asynchronous conversation that achieves even higher segmentation accuracy by combining lexical, conversational, and topic features.

Malioutov and Barzilay (2006) use a minimum cut clustering model to segment spoken lectures (i.e., spoken monolog). They form a weighted undirected graph where the nodes represent the sentences and the weighted edges represent the TF.IDF-based cosine similarity between the sentences. Then the segmentation can be solved as a graph partitioning problem with the assumption that the sentences in a segment should be similar, while sentences in different segments should be dissimilar. They optimize the normalized cut criterion (Shi & Malik, 2000) to extract the segments. In general, the minimization of the normalized cut is NP-complete. However, the sequentiality constraint of topic segmentation in monolog allows them to find an exact solution in polynomial time. Their approach performs better than the approach of Choi et al. (2001) in the corpus of spoken lectures. Since the sequentiality constraint does not hold in asynchronous conversation, we implement this model without this constraint by approximating the solution, and compare it with our models.

Probabilistic generative models, such as variants of Latent Dirichlet Allocation (LDA) (Blei, Ng, & Jordan, 2003) have also proven to be successful for topic segmentation in monolog and synchronous dialog. Blei and Moreno (2001) propose an aspect Hidden Markov Model (AHMM) to perform topic segmentation in written and spoken (i.e., transcribed) monologs, and show that the AHMM model outperforms the HMM for this task. Purver et al. (2006) propose a variant of LDA for segmenting meeting transcripts, and use the top words in the topic-word distributions as topic labels. However, their approach does not outperform LCSeg. Eisenstein and Barzilay (2008) propose another variant by incorporating cue words into the (sequential) segmentation model. In a follow-up work, Eisenstein (2009) proposes a constrained LDA model that uses *multi-scale lexical cohesion* to perform hierarchical topic segmentation. Nguyen, Boyd-Graber, and Resnik (2012) successfully incorporate speaker identity into a hierarchical nonparametric model for segmenting synchronous conversations (e.g., meeting, debate). In our work, we demonstrate how the general LDA model performs for topic segmentation in asynchronous conversation and propose a novel extension of LDA that exploits the fine conversational structure.

## 2.2 Topic Labeling

In the first comprehensive approach to topic labeling, Mei, Shen, and Zhai (2007) propose methods to label multinomial topic models (e.g., the topic-word distributions returned by LDA). Crucial to their approach is how they measure the semantic similarity between a topic-word distribution and a candidate topic label extracted from the same corpus. They perform such task by assuming another word distribution for the label and deriving the Kullback-Leibler divergence between the two distributions. It turns out that this measure is equivalent to the weighted point-wise mutual information (PMI) of the topic-words with the candidate label, where the weights are actually the probabilities in the topic-word distribution. They use Maximum Marginal Relevance (MMR) (Carbonell & Goldstein, 1998) to select the labels which are relevant, but not redundant. When labeling multiple topic-word distributions, to find discriminative labels, they adjust the semantic similarity scoring function such that a candidate label which is also similar to other topics gets lower score. In our work, we also use MMR to promote diversity in the labels for a topic. However, to get distinguishable labels for different topical segments in a conversation, we rank the words so that a high scoring word in one topic should not have high scores in other topics.

Recently, Lau, Grieser, Newman, and Baldwin (2011) propose methods to learn topic labels from Wikipedia titles. They use the top-10 words in each topic-word distribution to extract the candidate labels from Wikipedia. Then they extract a number of features to represent each candidate label. The features are actually different metrics used in the literature to measure the association between the topic words and the candidate label (e.g., PMI, t-test, chi-square test). They use Amazon Mechanical Turk to get humans to rank the top-10 candidate labels and use the average scores to learn a regression model.

Zhao, Jiang, He, Song, Achananuparp, Lim, and Li (2011a) addresses the problem of topical keyphrase extraction from Twitter. Initially they use a modified Twitter-LDA model (Zhao, Jiang, Weng, He, Lim, Yan, & Li, 2011b), which assumes a single topic assignment for a tweet, to discover the topics in the corpus. Then, they use a PageRank (Page, Brin, Motwani, & Winograd, 1999) to rank the words in each topic-word distribution. Finally, they perform a bi-gram test to generate keyphrases from the top ranked words in each topic.

While all the above studies try to mine topics from the whole corpus, our problem is to find the topical segments and label those for a given conversation, where topics are closely related and distributional variations are subtle (e.g., 'Game contents or size', 'Game design' in Figure 2). Therefore, statistical association metrics like PMI, t-test, chi-square test may not be reliable in our case because of data scarcity. Also at the conversation-level, the topics are too specific to a particular discussion (e.g., 'Telecon cancellation', 'TAG document', 'Responding to I18N' in Figure 1) that exploiting external knowledge bases like Wikipedia as a source of candidate labels is not a reasonable option for us. In fact, none of the human-authored labels in our developement set appears in Wikipedia as a title. Therefore, we propose to generate topic labels using a keyphrase extraction method that finds the most representative phrase(s) in the given text.

Several supervised and unsupervised methods have been proposed for keyphrase extraction (for a comprehensive overview see Medelyan, 2009). The supervised models (e.g., Hulth, 2003; Medelyan, Frnak, & Witten, 2009) follow the same two-stage framework. First, candidate keyphrases are extracted using n-gram sequences or a shallow parser (chunker). Second, a classifier filters the candidates. This strategy has been quite successful, but it is domain specific and labor intensive. Every new domain may require new annotations, which at times becomes too expensive and unrealistic. In contrast, our approach is to adopt an unsupervised paradigm, which is more robust across new domains, but still capable of achieving comparable performance to the supervised methods.

Mihalcea and Tarau (2004) use a graph-based (unsupervised) random walk model to extract keyphrases from journal abstracts and achieve the state-of-the-art performance (Mihalcea & Radev, 2011).[2] However, this model is generic and not designed to exploit properties of asynchronous conversations. We propose two novel random walk models to incorporate conversation specific information. Specifically, our models exploit information from two different sources: (i) from the leading sentences of the topical segments, and (ii) from the fine conversational structure of the conversation.

---

2. The original work was published before by Mihalcea and Tarau (2004).

## 2.3 Conversational Structure Extraction

Several approaches have been proposed to capture the underlying conversational structure of a conversation. Recent work on synchronous conversations has been focusing on disentangling multi-party chats, which have a linear structure. For example, several studies propose models to disentangle multi-party chat (Elsner & Charniak, 2010, 2011; Wang & Oard, 2009; Mayfield, Adamson, & Rosé, 2012). On the other hand, asynchronous conversations like email and social media services (e.g., Gmail, Twitter) generally organize comments into tree-structured threads using headers. Automatic methods to uncover such more complex structures have also been proposed (e.g., Wang, Wang, Zhai, & Han, 2011; Aumayr, Chan, & Hayes, 2011). However, the use of quotation in asynchronous conversations can express a conversational structure that is finer grained and can be more informative than the one revealed by reply-to relations between comments (Carenini et al., 2011). For example, in Figures 1 and 2, the proximity between a quoted paragraph and an unquoted one can represent an informative conversational link between the two (i.e., they talk about the same topic) that would not appear by only looking at the reply-to relations.

We previously presented a novel method to capture an email conversation at this finer level by analyzing the embedded quotations in emails (Carenini, Ng, & Zhou, 2007). A Fragment Quotation Graph (FQG) was formed, which was shown to be beneficial for email summarization (Carenini, Ng, & Zhou, 2008) and dialog act modeling (Joty, Carenini, & Lin, 2011). In this work, we generalize the FQG to any asynchronous conversation and demonstrate that topic segmentation and labeling models can also benefit significantly from this fine conversational structure of asynchronous conversation.

## 3. Topic Models for Asynchronous Conversations

Developing topic segmentation and labeling models for asynchronous conversations is challenging partly because of the specific characteristics of these media. As mentioned earlier, unlike monolog (e.g., a news article) and synchronous dialog (e.g., a meeting), topics in asynchronous conversations may not change in a sequential way, with topics being interleaved. Furthermore, as can be noticed in Figures 1 and 2, writing style varies among participants, and many people tend to use informal, short and ungrammatical sentences, thus making the discourse much less structured. One aspect of asynchronous conversation that at first glance may appear to help topic modeling is that each message comes with a header. However, often headers do not convey much topical information and sometimes they can even be misleading. For example, in the blog conversation (Figure 2), participants keep talking about different topics using the same title (i.e., 'nice nice nice'), which does not convey any topic information. Arguably, all these unique properties of asynchronous conversations limit the application of state-of-the-art techniques that have been successful in monolog and synchronous dialog. Below, we first describe these techniques and then we present how we have extended them to effectively deal with asynchronous conversations.

### 3.1 Topic Segmentation Models

We are the first to study the problem of topic segmentation in asynchronous conversation. Therefore, we first show how the existing models, which are originally developed for monolog and synchronous dialog, can be naively applied to asynchronous conversations. Then, by pointing out their limitations, we propose our novel topic segmentation models for asynchronous conversations.

#### 3.1.1 EXISTING MODELS

**LCSeg** (Galley et al., 2003) and **LDA** (Blei et al., 2003) are the two state-of-the-art unsupervised models for topic segmentation in monolog and synchronous dialog (Purver, 2011). In the following, we briefly describe these models and how they can be directly applied to asynchronous conversations.

*Lexical Cohesion-based Segmenter (LCSeg)*

LCSeg is a sequential segmentation model originally developed for segmenting meeting transcripts. It exploits the linguistic property called **lexical cohesion**, and assumes that topic changes are likely to occur where strong word repetitions start and end. It first computes **lexical chains** (Morris & Hirst, 1991) for each non-stop word based on word repetitions.[3] Then the chains are weighted according to their term frequency and the chain length. The more populated and compact chains get higher scores. The algorithm then works with two adjacent analysis windows, each of a fixed size $k$, which is empirically determined. At each sentence boundary, it computes the cosine similarity (or lexical cohesion function) between the two windows by representing each window as a vector of chain-scores of its words. Specifically, the lexical cohesion between windows ($X$ and $Y$) is computed with:

$$LexCoh(X, Y) = cos\_sim(X, Y) = \frac{\sum_{i=1}^{N} w_{i,X}.w_{i,Y}}{\sqrt{\sum_{i=1}^{N} w_{i,X}^2 \cdot \sum_{i=1}^{N} w_{i,Y}^2}} \qquad (1)$$

where $N$ is the number of chains and

$$w_{i,\Omega} = \begin{cases} rank(C_i) & \text{if chain } C_i \text{ overlaps } \Omega \in \{X, Y\} \\ 0 & \text{otherwise} \end{cases}$$

A sharp change at local minima in the resulting similarity curve signals a high probability of a topic boundary. The curve is smoothed, and for each local minimum it computes a segmentation probability based on its relative depth below its nearest peaks on either side. Points with the highest segmentation probability are then selected as hypothesized topic boundaries. This method is similar to TextTiling (Hearst, 1997) except that the similarity is computed based on the scores of the chains instead of term frequencies.

LCSeg can be directly applied to an asynchronous conversation by arranging its comments based on their arrival time (i.e., temporal order) and running the algorithm to get the topic boundaries.

---

3. One can also consider other lexical semantic relations (e.g., synonym, hypernym) in lexical chaining but the best results account for only repetition.

*Latent Dirichlet Allocation (LDA)*

LDA is a generative model that relies on the fundamental idea that documents are admixtures of topics, and a topic is a multinomial distribution over words. It specifies the following distribution over words within a document:

$$P(x_{ij}) = \sum_{k=1}^{K} P(x_{ij}|z_{ij} = k, \boldsymbol{b_k}) P(z_{ij} = k|\boldsymbol{\pi_i}) \tag{2}$$

where $K$ is the number of topics, $P(x_{ij}|z_{ij} = k, \boldsymbol{b_k})$ is the probability of word $x_{ij}$ in document $i$ for topic $k$, and $P(z_{ij} = k|\boldsymbol{\pi_i})$ is the probability that $k^{th}$ topic was sampled for the word token $x_{ij}$. We refer the multinomial distributions $\boldsymbol{b_k}$ and $\boldsymbol{\pi_i}$ as topic-word and document-topic distributions, respectively. Figure 3 shows the resultant graphical model in plate notation for $N$ documents, $K$ topics and $M_i$ tokens in each document $i$. Note that, $\alpha$ and $\beta$ are the standard Dirichlet priors on $\boldsymbol{\pi_i}$ and $\boldsymbol{b_k}$, respectively. Variational EM can be used to estimate $\boldsymbol{\pi}$ and $\boldsymbol{b}$ (Blei et al., 2003). One can also use Gibbs sampling to directly estimate the posterior distribution over $z$, i.e., $P(z_{ij} = k|x_{ij})$; namely, the topic assignments for word tokens (Steyvers & Griffiths, 2007).

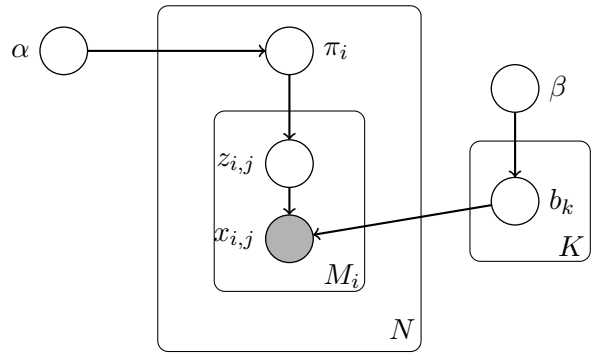

Figure 3: Graphical model for LDA in plate notation.

This framework can be directly applied to an asynchronous conversation by considering each comment as a document. By assuming the words in a sentence occur independently we can estimate the topic assignments for each sentence $s$ as follows:

$$P(z_m = k|s) = \prod_{x_m \in s} P(z_m = k|x_m) \tag{3}$$

Finally, the topic for $s$ can be assigned by:

$$k^* = argmax_k P(z_m = k|s) \tag{4}$$

### 3.1.2 Limitations of Existing Models

The main limitation of the two models discussed above is that they make the bag-of-words (BOW) assumption ignoring facts that are specific to a multi-party, asynchronous conversation. LCSeg considers only term frequency and how closely these terms occur in the temporal order of the sentences. If topics are interleaved and do not change sequentially in the temporal order, as is often the case in an asynchronous conversation, then LCSeg would fail to find the topical segments correctly.

On the other hand, the only information relevant to LDA is term frequency. Several extensions of LDA over the BOW approach have been proposed. For example, Wallach (2006) extends the model beyond BOW by considering n-gram sequences. Griffiths, Steyvers, Blei, and Tenenbaum (2005) present an extension that is sensitive to word-order and automatically learns the syntactic as well as semantic factors that guide word choice. Boyd-Graber and Blei (2008) describe another extension to consider the syntax of the sentences.

We argue that these models are still inadequate for finding topical segments correctly in asynchronous conversations especially when topics are closely related and their distributional variations are subtle (e.g., 'Game contents or size' and 'Game design'). To better identify the topics one needs to consider the features specific to asynchronous conversations (e.g., conversation structure, speaker, recipient). In the following, we propose our novel unsupervised and supervised topic segmentation models that incorporate these features.

### 3.1.3 PROPOSED UNSUPERVISED MODELS

One of the most important indicators for topic segmentation in asynchronous conversation is its conversation structure. As can be seen in the examples (Figures 1 and 2), participants often reply to a post and/or use quotations to talk about the same topic. Notice also that the use of quotations can express a conversational structure that is at a finer level of granularity than the one revealed by reply-to relations. In our corpora, we found an average quotation usage of 9.85 per blog conversation and 6.44 per email conversation. Therefore, we need to leverage this key information to get the best out of our models. Specifically, we need to capture the conversation structure at the quotation (i.e., text fragment) level, and to incorporate this structure into our segmentation models in a principled way.

In the following, we first describe how we can capture the conversation structure at the fragment level. Then we show how the unsupervised models LCSeg and LDA can be extended to take this conversation structure into account, generating two novel unsupervised models for topic segmentation in asynchronous conversation.

*Extracting Finer-level Conversation Structure*

Since consecutive turns in asynchronous conversations can be far apart in time, when participants reply to a post or comment, a quoted version of the original message is often included (specially in email) by default in the draft reply in order to preserve context. Furthermore, people tend to break down the quoted message so that different questions, requests or claims can be dealt with separately. As a result, each message, unless it is at the beginning, will contain a mix of quoted and novel paragraphs (or fragments) that may well reflect a reply-to relationship between paragraphs that is at a finer level of granularity than the one explicitly recorded between comments. We proposed a novel approach to capture this finer level conversation structure in the form of a graph called **Fragment Quotation Graph (FQG)** (Carenini et al., 2007). In the following, we demonstrate how to build a FQG for the sample blog conversation shown in Figure 2. Figure 4(a) shows the same blog conversation, but for the sake of illustration, instead of showing the real content, we abbreviate it as a sequence of labels (e.g., $a, b$), each label corresponding to a text fragment (see the Fragment column in Figure 2). Building a FQG is a two-step process.

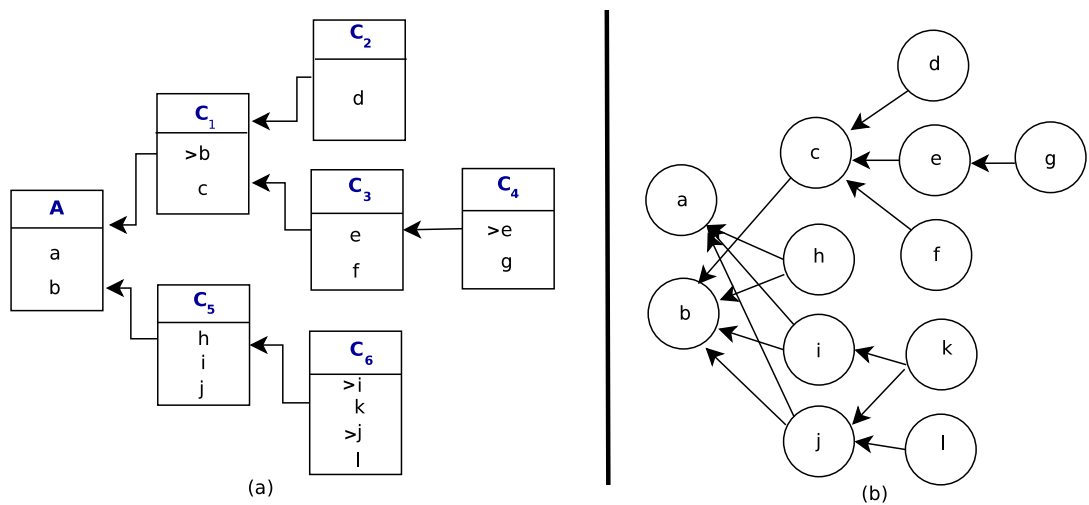

Figure 4: (a) The main $A$rticle and the $C$omments with the fragments for the example in Figure 2. Arrows indicate 'reply-to' relations. (b) The corresponding FQG.

- *Node creation:* Initially, by processing the whole conversation, we identify the new and quoted fragments of different depth levels. The depth level of a quoted fragment is determined by the number of quotation marks (e.g., $>$, $>>$, $>>>$). For instance, comment $C_1$ contains a new fragment $c$ and a quoted fragment $b$ of depth level 1. $C_6$ contains two new fragments $k$ and $l$, and two quoted fragments $i$ and $j$ of depth level 1, and so on. Then in the second step, we compare the fragments with each other and based on their lexical overlap we find the distinct fragments. If necessary, we split the fragments in this step. For example, $ef$ in $C_3$ is divided into $e$ and $f$ distinct fragments when compared with the fragments of $C_4$. This process gives 12 distinct fragments which constitute the nodes of the FQG shown in Figure 4(b).

- *Edge creation:* We create edges to represent likely replying relationship between fragments assuming that any new fragment is a potential reply to its neighboring quotations of depth level 1. For example, for the fragments of $C_6$ in Figure 4(a), we create two edges from $k$ (i.e., (k,i),(k,j)) and one edge from $l$ (i.e., (l,j)) in Figure 4(b). If a comment does not contain any quotation, then its fragments are linked to the new fragments of the comment to which it replies, capturing the original 'reply-to' relation.

Note that the FQG is only an approximation of the reply relations between fragments. In some cases, proximity may not indicate any connection and in other cases a connection can exist between fragments that are never adjacent in any comment. Furthermore, this process could lead to less accurate conversational structure when quotation marks (or cues) are not present. Nonetheless, we previously showed that considering the FQG can be beneficial in dialog act modeling (Joty et al., 2011) and email summarization (Carenini et al., 2008). In this study, we show that topic segmentation (this Section) and labeling (Section 3.2) models can also benefit significantly from this fine conversational structure of asynchronous conversation. Minimizing the noise in FQGs is left as future work.

### LCSeg with FQG (LCSeg+FQG)

If we examine the FQG carefully, the paths (considering the fragments of the first comment as root nodes) can be interpreted as subconversations, and topic shifts are likely to occur along the pathway as we walk down a path. We incorporate FQG into LCSeg in three steps.

- *Path extraction:* First, we extract all the paths of a FQG. For example, for the FQG in Figure 4(b), we extract the paths $< a, j, l >$, $< b, c, e, g >$, $< b, c, d >$, and so on.

- *LCSeg application:* We then run LCSeg algorithm on each of the extracted paths separately and collect the segmentations. For example, when applied LCSeg to $< b, c, e, g >$ and $< b, c, d >$ paths separately, we may get the following segmentations- $< b, c \mid e, g >$ and $< b, c \mid d >$, where '|' denotes the segment boundary.[4] Notice that a fragment can be in multiple paths (e.g., $b$, $c$) which will eventually cause its sentences to be in multiple segments. So, in the final step, we need a consolidation method.

- *Consolidation:* Our intuition is that sentences in a consolidated segment should appear together in a segment more often when LCSeg is applied in step 2, and if they do not appear together in any segment, they should at least be similar. To achieve this, we construct a weighted undirected graph $G(V, E)$, where the nodes $V$ represent the sentences and the edge weights $w(x, y)$ represent the number of segments in which sentences $x$ and $y$ appear together; if $x$ and $y$ do not appear together in any segment, then their cosine similarity is used as edge weights. More formally,

$$w(x, y) = \begin{cases} n, & \text{if } x \text{ and } y \text{ appear together in } n \text{ segments and } n > 0 \\ cos\_sim(x, y), & \text{if } n = 0 \end{cases}$$

We measure the cosine similarity between sentences $x$ and $y$ as follows:

$$cos\_sim(x, y) = \frac{\sum_{w \in x, y} tf_{w,x} \cdot tf_{w,y}}{\sqrt{\sum_{x_i \in x} tf_{x_i, x}^2} \cdot \sqrt{\sum_{y_i \in y} tf_{y_i, y}^2}} \tag{5}$$

where $tf_{a,s}$ denotes the term frequency of term $a$ in sentence $s$. The cosine similarity ($0 \leq cos\_sim(x, y) \leq 1$) provides informative edge weights for the sentence pairs that are not directly connected by LCSeg segmentation decisions.[5] Now, the consolidation problem can be formulated as a **k-way-mincut** graph partitioning problem with the **normalized cut (Ncut)** criterion (Shi & Malik, 2000):

---

4. For convenience, we are showing the segmentations at the fragment level, but the segmentations are actually at the sentence level.

5. In our earlier work (Joty, Carenini, Murray, & Ng, 2010), we did not consider the cosine similarity when two sentences do not appear together in any of the segments. However, later we found out that including the cosine similarity offers more than 2% absolute gain in segmentation performance.

$$Ncut_k(V) = \frac{cut(A_1, V - A_1)}{assoc(A_1, V)} + \frac{cut(A_2, V - A_2)}{assoc(A_2, V)} + \cdots + \frac{cut(A_k, V - A_k)}{assoc(A_k, V)} \quad (6)$$

where $A_1, A_2 \cdots A_k$ form a partition (i.e., disjoint sets of nodes) of the graph, and $V - A_k$ is the set difference between $V$ (i.e., set of all nodes) and $A_k$. The $cut(A, B)$ measures the total edge weight from the nodes in set $A$ to the nodes in set $B$, and $assoc(A, V)$ measures the total edge weight from the nodes in set $A$ to all nodes in the graph. More formally:

$$cut(A, B) = \sum_{u \in A, v \in B} w(u, v) \quad (7)$$

$$assoc(A, V) = \sum_{u \in A, t \in V} w(u, t) \quad (8)$$

Note that the partitioning problem can be solved using any correlation clustering method (e.g., Bansal, Blum, & Chawla, 2002). Previous work on graph-based topic segmentation (Malioutov & Barzilay, 2006) has shown that the Ncut criterion is more appropriate than just the cut criterion, which accounts only for total edge weight connecting $A$ and $B$, and therefore, favors cutting small sets of isolated nodes in the graph. However, solving Ncut is NP-complete. Hence, we approximate the solution following the method proposed by Shi and Malik, (2000), which is time efficient and has been successfully applied to image segmentation in computer vision.

Notice that this approach makes a difference only if the FQG of the conversation contains more than one path. In fact, in our corpora we found an average number of paths of 7.12 and 16.43 per email and blog conversations, respectively.

### LDA with FQG (LDA+FQG)

A key advantage of probabilistic Bayesian models, such as LDA, is that they allow us to incorporate multiple knowledge sources in a coherent way in the form of priors (or regularizer). To incorporate FQG into LDA, we propose to regularize LDA so that two sentences in the same or adjacent fragments are likely to appear in the same topical cluster. The first step towards this aim is to regularize the topic-word distributions (i.e., $\boldsymbol{b}$ in Figure 3) with a **word network** such that two connected words get similar topic distributions.

For now, assume that we are given a word network as an undirected graph $G(V, E)$, with nodes $V$ representing the words and the edges $(u, v) \in E$ representing the links between words $u$ and $v$. We want to regularize the topic-word distributions of LDA such that two connected words $u$ and $v$ in the word network have similar topic distributions (i.e., $\boldsymbol{b_k}^{(u)} \approx \boldsymbol{b_k}^{(v)}$ for $k = 1 \ldots K$). The standard conjugate Dirichlet prior $\mathrm{Dir}(\boldsymbol{b_k}|\beta)$, however does not allow us to do that, because here all words share a common variance parameter, and are mutually independent except normalization constraint (Minka, 1999). Recently, Andrzejewski, Zhu, and Craven (2009) describe a method to encode **must-links** and **cannot-links** between words using a Dirichlet Forest prior. Our goal is just to encode the must-links. Therefore, we reimplemented their model with its capability of encoding just the (must-)links.

Must-links between words such as $(a, b), (b, c)$, or $(x, y)$ in Figure 5(a) can be encoded into LDA using a **Dirichlet Tree (DT)** prior. Like the traditional Dirichlet, DT prior is also a conjugate to the multinomial, but under a different parameterization. Instead of representing a multinomial sample as the outcome of a K-sided die, in the tree representation (e.g., Figure 5(b)), a sample (i.e., a leaf in the tree) is represented as the outcome of a finite stochastic process. The probability of a leaf (i.e., a word in our case) is the product of branch probabilities leading to that leaf. A DT prior is the distribution over leaf probabilities.

Let $\omega^n$ be the edge weight leading into node $n$, $C(n)$ be the children of node $n$, $L$ be the leaves of the tree, $I$ be the internal nodes, and $L(n)$ be the leaves in the subtree under node $n$. We generate a sample $\boldsymbol{b_k}$ from $DT(\omega)$ by drawing a multinomial at each internal node $i \in I$ from $\text{Dir}(\omega^{C(i)})$ (i.e., the edge weights from node $i$ to its children). The probability density function of $DT(\boldsymbol{b_k}|\omega)$ is given by:

$$DT(\boldsymbol{b_k}|\omega) \propto \left( \prod_{l \in L} b_l^{k^{\omega^l - 1}} \right) \left( \prod_{i \in I} \left( \sum_{j \in L(i)} b_j^k \right)^{\Delta(i)} \right) \tag{9}$$

where $\Delta(i) = \omega^i - \sum_{j \in C(i)} \omega^j$, the difference between the in-degree and out-degree of an internal node $i$. Notice when $\Delta(i) = 0$ for all $i \in I$, the DT reduces to the standard Dirichlet.

Suppose we are given the word network as shown in Figure 5(a). The network can be decomposed into a collection of chains (e.g., $(a, b, c)$, $(p)$, and $(x, y)$). For each chain containing multiple elements (e.g., $(a, b, c), (x, y)$), there is a subtree in the DT (Figure 5(b)), with one internal node (blank in Figure) and the words of the chain as its leaves. The weight from the internal node to each of its leaves is $\lambda\beta$, where $\lambda$ is the regularization strength and $\beta$ is the parameter of the standard symmetric Dirichlet prior on $\boldsymbol{b_k}$. The root node of the DT then connects to the internal nodes with $|L(i)|\beta$ weight. The leaves (words) for the single element chains (e.g, $(p)$) are then connected to the root of the DT directly with weight $\beta$. Notice that when $\lambda = 1$, $\Delta(i) = 0$, and it reduces to the standard LDA (i.e., no regularization). By tuning $\lambda$ we control the strength of the regularization.

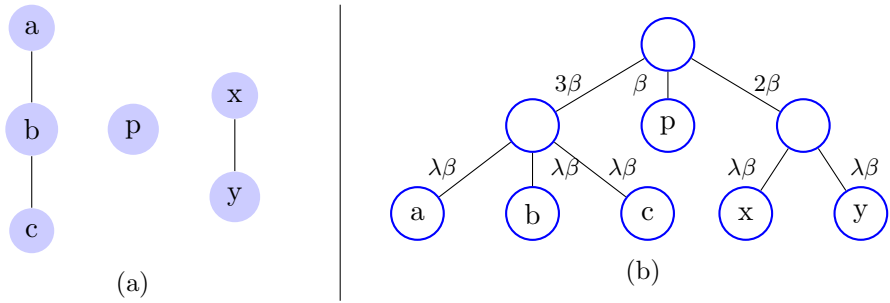

(a)              (b)

Figure 5: (a) Sample word network, (b) A Dirichlet Tree (DT) built from such word network.

At this point what is left to be explained is how we construct the word network. To regularize LDA with a FQG, we construct the word network where a word is linked to the words in the same or adjacent fragments in the FQG. Specifically, if word $w_i \in frag_x$ and word $w_j \in frag_y$ and $w_i \neq w_j$, we create a link $(w_i, w_j)$ if $x = y$ or $(x, y) \in E_{fqg}$, where $E_{fqg}$ is the set of edges in the FQG. This implicitly compels two sentences in the same or adjacent fragments to have similar topic distributions, and appear in the same topical segment.

### 3.1.4 Proposed Supervised Model

Although the unsupervised models discussed in the previous section have the key advantage of not requiring any labeled data, they can be limited in their ability to learn domain-specific knowledge from a possible large and diverse set of features (Eisenstein & Barzilay, 2008). Beside discourse cohesion, which captures the changes in content, there are other important domain-specific distinctive features which signal topic change. For example, discourse markers (or cue phrases) (e.g., *okay, anyway, now, so*) and prosodic cues (e.g., longer pause) directly provide clues about topic change, and have been shown to be useful features for topic segmentation in monolog and synchronous dialog (Passonneau & Litman, 1997; Galley et al., 2003). We hypothesize that asynchronous conversations can also feature their own distinctive characteristics for topic shifts. For example, features like *sender* and *recipient* are arguably useful for segmenting asynchronous conversations, as different participants can be more or less active during the discussion of different topics. Therefore, as a next step to build an even more accurate topic segmentation model for asynchronous conversations, we propose to combine different sources of possibly useful information in a principled way.

The supervised framework serves as a viable option to combine a large number of features and optimize their relative weights for decision making, but relies on labeled data for training. The amount of labeled data required to achieve an acceptable performance is always an important factor to consider for choosing supervised vs. unsupervised. In this work, we propose a supervised segmentation model that outperforms all the unsupervised models, even when it is trained on a small number of labelled conversations.

Our supervised model is built on the **graph-theoretic** framework which has been used in many NLP tasks, including coreference resolution (Soon, Ng, & Lim, 2001) and chat disentanglement (Elsner & Charniak, 2010). This method works in two steps.

- *Classification:* A binary classifier which is trained on a labeled dataset, marks each pair of sentences of a given conversation as **same** or **different** topics.

- *Graph partitioning:* A weighted undirected graph $G = (V, E)$ is formed, where the nodes $V$ represent the sentences in the conversation and the edge-weights $w(x, y)$ denote the probability (given by the classifier) of the two sentences $x$ and $y$ to appear in the same topic. Then an optimal partition is extracted.

*Sentence pair classification*

The classifier's accuracy in deciding whether a pair of sentences $x$ and $y$ is in the **same** or **different** topics is crucial for the model's performance. Note that since each sentence pair of a conversation defines a data point, a conversation containing $n$ sentences gives $1 + 2 + \ldots + (n\text{-}1) = \frac{n(n-1)}{2} = O(n^2)$ training examples. Therefore, a training dataset containing $m$ conversations gives $\sum_{i=1}^{m} \frac{n_i(n_i-1)}{2}$ training examples, where $n_i$ is the number of sentences in the $i^{th}$ conversation. This quadratic expansion of training examples enables the classifier to achieve its best classification accuracy with only a few labeled conversations.

By pairing up the sentences of each email conversation in our email corpus, we got a total of $14,528$ data points of which $58.8\%$ are in the *same* class (i.e., *same* is the most likely in email), and by pairing up the sentences of each blog conversation in our blog corpus, we got a total of $572,772$ data points of which $86.3\%$ are in the *different* class (i.e., *different* is the most likely in blog).[6] To select the best classifier, we experimented with a variety of classifiers with the full feature set (Table 2). Table 1 shows the performance of the classifiers averaged over a **leave-one-out** procedure, i.e., for a corpus containing $m$ conversations, train on $m-1$ conversations and test on the rest.

| Classifier | Type | Regularizer | Accuracy (**Blog**) | | Accuracy (**Email**) | |
|---|---|---|---|---|---|---|
| | | | Train | Test | Train | Test |
| KNN | non-parametric | - | 62.7% | 61.4 % | 54.6% | 55.2% |
| **LR** | **parametric** | $l_2$ | **90.8%** | **91.9%** | **71.7%** | **72.5%** |
| LR | parametric | $l_1$ | 86.8% | 87.6% | 69.9% | 67.7% |
| RMLR (rbf) | non-parametric | $l_2$ | 91.7% | 82.0% | 91.1% | 62.1% |
| SVM (lin) | parametric | - | 76.6% | 78.7 % | 68.3% | 69.6% |
| SVM (rbf) | non-parametric | - | 80.5% | 77.9% | 75.9% | 67.7% |
| Majority class | - | - | 86.3% (different) | | 58.8% (same) | |

Table 1: Performance of the classifiers using the full feature set (Table 2). For each training set, regularizer strength $\lambda$ (or C in SVMs) was learned by 10-fold cross validation.

K-Nearest Neighbor (KNN) performs very poorly. Logistic Regression (LR) with $l_2$ regularizer delivers the highest accuracy on both datasets. Support Vector Machines (SVMs) (Cortes & Vapnik, 1995) with linear and rbf kernels perform reasonably well, but not as well as LR. The Ridged Multinomial Logistic Regression (RMLR) (Krishnapuram et al. 2005), a kernelized LR, extremely overfits the data. We opted for the LR with $l_2$ regularizer because it not only delivers the best performance in term of accuracy, but it is also very efficient. The limited memory BFGS (L-BFGS) fitting algorithm used in LR is efficient in terms of both time (quadratic convergence rate; fastest among the listed models) and space ($O(mD)$, where $m$ is the memory parameter of L-BFGS and $D$ is the number of features).

Table 2 summarizes the full feature set and the mean test set accuracy (using leave-one-out procedure) achieved with different types of features in our LR classifier.

**Lexical** features encode similarity between two sentences $x$ and $y$ based on their raw content. Term frequency-based similarity is a widely used feature in previous work, e.g., TextTiling (Hearst, 1997). We compute this feature by considering two analysis windows, each of fixed size $k$. Let $X$ be the window including sentence $x$ and the preceding $k-1$ sentences, and $Y$ be the window including sentence $y$ and the following $k-1$ sentences. We measure the cosine similarity between the two windows by representing them as vectors of **TF.IDF** (Salton & McGill, 1986) values of the words. Another important domain specific feature that proved to be useful in previous research (e.g., Galley et al., 2003) is **cue words** (or discourse markers) that sign the presence of a topic boundary (e.g., 'coming up', 'joining us' in news). Since our work concerns conversations (not monologs), we adopt the cue word list derived automatically from a meeting corpus by Galley et al. (2003). If $y$ answers or greets $x$ then it is likely that they are in the same topic. Therefore, we use the Question Answer (**QA**) pairs and **greeting** words as two other lexical features.

---

6. See Section 4 for a detailed description of our corpora. The class labels are produced by taking the maximum vote of the three annotators.

| **Lexical** | **Accuracy: 86.8 Precision: 62.4 Recall: 4.6 (Blog)** |
| | **Accuracy: 59.6 Precision: 59.7 Recall: 99.8 (Email)** |

| $TFIDF_1$ | TF.IDF-based similarity between $x$ and $y$ with window size k=1. |
| $TFIDF_2$ | TF.IDF-based similarity between $x$ and $y$ with window size k=2. |
| Cue Words | Either $x$ or $y$ contains a cue word. |
| QA | $x$ asks a question explicitly using ? and $y$ answers it using any of (yes, yeah, okay, ok, no, nope). |
| Greet | Either $x$ or $y$ has a greeting word (hi, hello, thanks, thx, tnx, thank.) |

| **Conversation** | **Accuracy: 88.2 Precision: 81.6 Recall: 20.5 (Blog)** |
| | **Accuracy: 65.3 Precision: 66.7 Recall: 85.1 (Email)** |

| Gap | The gap between $y$ and $x$ in number of sentence(s). |
| Speaker | $x$ and $y$ have the same sender (yes or no). |
| $FQG_1$ | Distance between $x$ and $y$ in FQG in terms of fragment id. (i.e., $|frag\_id(y) - frag\_id(x)|$). |
| $FQG_2$ | Distance between $x$ and $y$ in FQG in terms of number of edges. |
| $FQG_3$ | Distance between $x$ and $y$ in FQG in number of edges but this time considering it as an undirected graph. |
| Same/Reply | whether $x$ and $y$ are in the same comment or one is a reply to the other. |
| Name | $x$ mentions $y$'s speaker or vice versa. |

| **Topic** | **Accuracy: 89.3 Precision: 86.4 Recall: 17.3 (Blog)** |
| | **Accuracy: 67.5 Precision: 68.9 Recall: 76.8 (Email)** |

| $LSA_1$ | LSA-based similarity between $x$ and $y$ with window size k=1. |
| $LSA_2$ | LSA-based similarity between $x$ and $y$ with window size k=2. |
| LDA | LDA segmentation decision on $x$ and $y$ (same or different). |
| LDA+FQG | LDA+FQG segmentation decision on $x$ and $y$ (same or different). |
| LCSeg | LCSeg segmentation decision on $x$ and $y$ (same or different). |
| LCSeg+FQG | LCSeg+FQG segmentation decision on $x$ and $y$ (same or different). |
| LexCoh | Lexical cohesion between $x$ and $y$. |

| **Combined** | **Accuracy: 91.9 Precision: 78.8 Recall: 25.8 (Blog)** |
| | **Accuracy: 72.5 Precision: 70.4 Recall: 81.5 (Email)** |

Table 2: Features with average performance on testsets (using leave-one-out).

**Conversational** features capture conversational properties of an asynchronous conversation. Time gap and speaker are commonly used features for segmenting synchronous conversations (e.g., Galley et al. 2003). We encode similar information in asynchronous media by counting the number of sentences between $x$ and $y$ (in their temporal order) as the **gap**, and their senders as the **speakers**. The strongest baseline Speaker (see Section 5.1) also proves its effectiveness in asynchronous domains. The results in Section 5.1 also suggest that fine conversational structure in the form of FQG can be beneficial when it is incorporated into the unsupervised segmentation models. We encode this valuable information into our supervised segmentation model by computing three distance features on the FQG: $\boldsymbol{FQG_1}, \boldsymbol{FQG_2}, \boldsymbol{FQG_3}$. State-of-the-art email and blog systems use reply-to relation to group comments into threads. If $y$'s comment is **same as or reply to** $x$'s comment, then it is likely that the two sentences talk about the same topic. Participants sometimes mention each other's **name** in multi-party conversations to make disentanglement easier (Elsner & Charniak, 2010). We also use this as a feature in our supervised segmentation model.

**Topic** features are complex and encode topic information from existing segmentation models. Choi et al. (2001) used Latent Semantic Analysis (**LSA**) to measure the similarity between two sentences and showed that the LSA-based similarity yields better results than the direct TF.IDF-based similarity since it surmounts the problems of synonymy (e.g., car, auto) and polysemy (e.g., money bank, river bank). To compute LSA, we first construct a word-document matrix $W$ for a conversation, where $W_{i,j}$ = the frequency of word $i$ in comment $j$ × the IDF score of word $i$. We perform truncated Singular Value Decomposition (SVD) of $W$: $W \approx U_k \Sigma_k V_k^T$, and represent each word $i$ as a $k$ dimensional[7] vector $\boldsymbol{\Lambda}_i^k$. Each sentence is then represented by the weighted sum of its word vectors. Formally, the LSA representation for sentence $s$ is $\boldsymbol{\Lambda}_s = \sum_{i \in s} tf_i^s . \boldsymbol{\Lambda}_i^k$, where $tf_i^s$ = the term frequency of word $i$ in sentence $s$. Then just like the TF.IDF-based similarity, we compute the LSA-based similarity between sentences $x$ and $y$, but this time by representing the corresponding windows (i.e., $X$ and $Y$) as LSA vectors.

The segmentation decisions of **LDA**, **LDA+FQG**, **LCSeg** and **LCSeg+FQG** models described in the previous section are also encoded as topic features.[8] As described in Section 3.1.1, LCSeg computes a lexical cohesion (**LexCoh**) function between two consecutive windows based on the scores of the lexical chains. Galley et al. (2003) shows a significant improvement when this function is used as a feature in the supervised (sequential) topic segmentation model for meetings. However, since our problem of topic segmentation is not sequential, we want to compute this function for any two given windows $X$ and $Y$ (not necessary consecutive). To do that, we first extract the lexical chains with their scores and spans (i.e., beginning and end sentence numbers) for the conversation. The lexical cohesion function is then computed with the method described in Equation 1.

---

7. In our study, $k$ was empirically set to $\frac{1}{4}$×number of comments based on a held-out development set.
8. Our earlier work (Joty, Carenini, Murray, & Ng, 2011) did not include the segmentation decisions of LDA+FQG and LCSeg+FQG models as features. However, including these features improves both classification accuracy and segmentation accuracy.

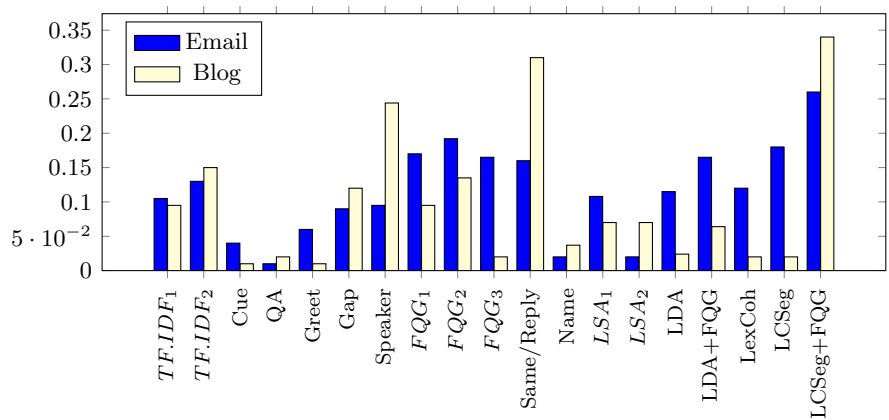

Figure 6: Relative importance of the features averaged over leave-one-out.

We describe our classifier's performance in terms of raw accuracy (correct decisions/total), precision and recall of the *same* class for different types of features averaged over a leave-one-out procedure (Table 2). Among the feature types, topic features yield the highest accuracy and same-class precision in both corpora ($p < 0.01$).[9] Conversational features also have proved to be important and achieve higher accuracy than lexical features ($p < 0.01$). Lexical features have poor accuracy, only slightly higher than the majority baseline that always picks the most likely class. However, when we combine all the features, we get the best performance ($p < 0.005$). These results demonstrate the importance of topical and conversational features beyond the lexical features used only by the existing segmentation models. When we compare the performance on the two corpora, we notice that while in blog the accuracy and the same-class precision are higher than in email, the same-class recall is much lower. Although this is reasonable given the class distributions in the two corpora (i.e., 13.7% and 58.8% examples are in the same-class in blog and email, respectively), surprisingly, when we tried to deal with this problem by applying the bagging technique (Breiman, 1996), the performance does not improve significantly. Note that some of the classification errors occurred in the sentence-pair classification phase are recovered in the graph partitioning step (see below). The reason is that the incorrect decisions will be outvoted by the nearby sentences that are clustered correctly.

We further analyze the contribution of individual features. Figure 6 shows the relative importance of the features based on the absolute values of their coefficients in our LR classifier. The segmentation decision of LCSeg+FQG is the most important feature in both domains. The Same/Reply is also an effective feature, especially in blog. In blog, the Speaker feature also plays an important role. The $FQG_2$ (distance in number of edges in the directed FQG) is also effective in both domains, especially in email. The other two features on FQG ($FQG_1$, $FQG_3$) are also very relevant in email.

Finally, in order to determine how many annotated conversations we need to achieve the best segmentation performance, Figure 7 shows the classification error rate (incorrect decisions/total), tested on 5 randomly selected conversations and trained on an increasing number of randomly added conversations. Our classifier appears to achieve its best performance with a small number of labeled conversations. For blog, the error rate flattens with only 8 conversations, while for email, this happens with about 15. This is not surprising since blog conversations are much longer (an average of 220.55 sentences) than email conversations (an average of 26.3 sentences), generating a similar number of training examples with only a few conversations (recall, for $n$ sentences we get $O(n^2)$ training examples).

---

9. All tests of statistical significance were performed using paired t-test.

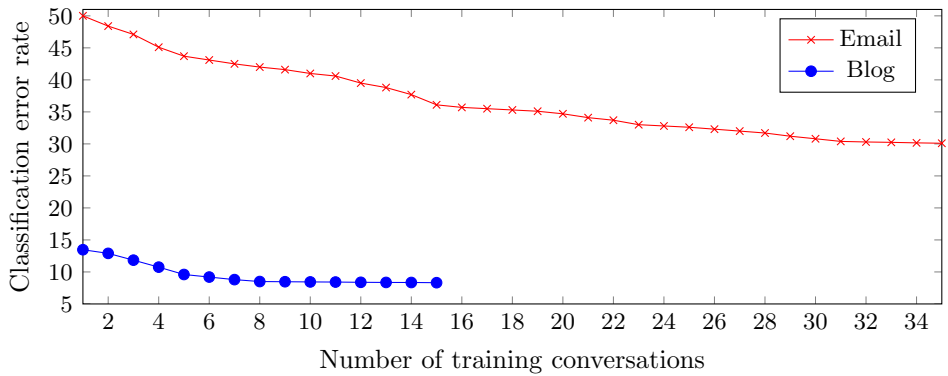

Figure 7: Error rate vs. number of training conversations.

*Graph partitioning*

Given a weighted undirected graph $G = (V, E)$, where the nodes $V$ represent the sentences and the edge-weights $w(x, y)$ denote the probability (given by our classifier) of the two sentences $x$ and $y$ to appear in the same topic, we again formulate the segmentation task as a **k-way-mincut** graph partitioning problem with the intuition that sentences in a segment should discuss the same topic, while sentences in different segments should discuss different topics. We optimize the **normalized cut** criterion (i.e., equation 6) to extract an optimal partition as was done before for consolidating various segments in LCSeg+FQG.

## 3.2 Topic Labeling Models

Now that we have methods to automatically identify the topical segments in an asynchronous conversation, the next step in the pipeline is to generate one or more informative descriptions or topic labels for each segment to facilitate interpretations of the topics. We are the first to address this problem in asynchronous conversation.

Ideally, a topic label should be meaningful, semantically similar to the underlying topic, general and discriminative (when there are multiple topics) (Mei et al., 2007). Traditionally, the top $k$ words in a multinomial topic model like LDA are used to describe a topic. However, as pointed out by Mei et al., at the word-level, topic labels may become too generic and impose cognitive difficulties on a user to interpret the meaning of the topic by associating the words together. For example, in Figure 2, without reading the text, from the words {release, free, reaction, Daggerfall}, it may be very difficult for a user to understand that the topic is about Daggerfall's free release and people's reaction to it. On the other hand, if the labels are expressed at the sentence-level, they may become too specific to cover the whole theme of the topic (Mei et al., 2007). Based on these observations, recent studies (e.g., Mei et al., 2007; Lau et al., 2011) advocate for **phrase-level** topic labels, which are also consistent with the monolog corpora built as a part of the Topic Detection and Tracking (TDT) project[10]. Note that we also observe a preference for phrase-level labels within our own asynchronous conversational corpora in which human annotators without specific instructions spontaneously generated topic labels at the phrase-level. Considering all this, we treat phrase-level as our target level of granularity for a topic label.

Our problem is no different from the problem of **keyphrase indexing** (Medelyan, 2009) where the task is to find a set of keyphrases either from the given text or from a controlled vocabulary (i.e., domain-specific terminologies) to describe the topics covered in the text. In our setting, we do not have such a controlled vocabulary. Furthermore, exploiting generic knowledge bases like Wikipedia as a source of devising such a controlled vocabulary (Medelyan, 2009) is not a viable option in our case since the topics are very specific to a particular discussion (e.g., *Free release of Daggerfall and reaction*, *Game contents or size* in Figure 2). In fact, none of the human-authored labels in our developement set appears verbatim in Wikipedia. We propose to generate topic labels using a **keyphrase extraction** approach that identifies the most representative phrase(s) in the given text. We adapt a graph-based **unsupervised** ranking framework, which is domain independent, and without relying on any labeled data achieves state-of-the-art performance on keyphrase extraction (Mihalcea & Radev, 2011). Figure 8 shows our topic labeling framework. Given a (topically) segmented conversation, our system generates $k$ keyphrases to describe each topic in the conversation. Below we discuss the different components of the system.

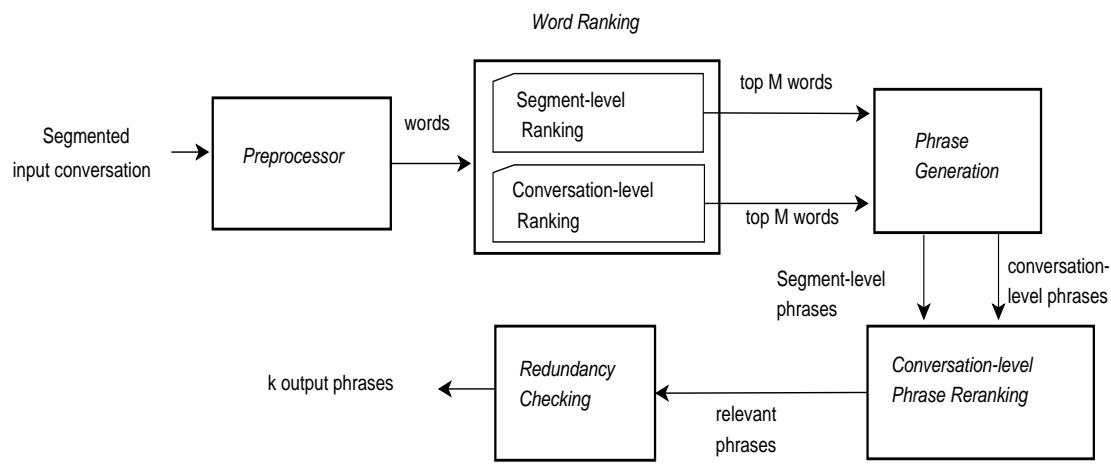

Figure 8: Topic labeling framework for asynchronous conversation.

### 3.2.1 Preprocessing

In the preprocessing step, we tokenize the text and apply a **syntactic filter** to select the words of a certain part-of-speech (POS). We use a state-of-the-art tagger[11] to tokenize the text and annotate the tokens with their POS tags. We experimented with five different syntactic filters. They select (i) nouns, (ii) nouns and adjectives, (iii) nouns, adjectives and verbs, (iv) nouns, adjectives, verbs and adverbs, and (v) all words, respectively. The filters also exclude the stopwords. The second filter, that selects only nouns and adjectives, achieves the best performance on our development set, which is also consistent with the finding of Mihalcea and Tarau (2004). Therefore, this syntactic filter is used in our system.

---

10. http://projects.ldc.upenn.edu/TDT/
11. Available at http://cogcomp.cs.illinois.edu/page/software

### 3.2.2 WORD RANKING

The words selected in the preprocessing step correspond to the nodes in the graph. A direct application of the ranking method described by Mihalcea and Tarau (2004) would define the edges based on the **co-occurrence** relation between the respective words, and then apply the PageRank (Page et al., 1999) algorithm to rank the nodes. We argue that co-occurrence relations may be insufficient for finding topic labels in an asynchronous conversation. To better identify the labels one needs to consider aspects that are specific to asynchronous conversation. In particular, we propose to exploit two different forms of conversation specific information into our graph-based ranking model: (1) informative clues from the **leading sentences** of a topical segment, and (2) the fine-grained conversational structure (i.e., the **Fragment Quotation Graph (FQG)**) of an asynchronous conversation. In the following, we describe these two novel extensions in turn.

#### *Incorporating Information from the Leading Sentences*

In general, the leading sentences of a topic segment carry informative clues for the topic labels, since this is where the speakers will most likely try to signal a topic shift and introduce the new topic. Our key observation is that this is especially true for asynchronous conversations, in which topics are interleaved and less structured. For example, in Figure 2, notice that in almost every case, the leading sentences of the topical segments covers the information conveyed by the labels. This property is further confirmed in Figure 9, which shows the percentage of non-stopwords in the human-authored labels that appear in leading sentences of the segments in our development set. The first sentence covers about 29% and 38% of the words in the gold labels in the blog and email corpora, respectively. The first two sentences cover around 35% and 45% of the words in the gold labels in blog and email, respectively. When we consider the first three sentences, the coverage raises up to 39% and 49% for blog and email, respectively. The increment is less as we add more sentences.

To leverage this useful information in our ranking model, we propose the following **biased random walk** model, where $P(w|U_k)$, the score of a word $w$ given a set of leading sentences $U_k$ in topic segment $k$, is expressed as a convex combination of its relevance to the leading sentences $U_k$ (i.e., $\rho(w|U_k)$) and its relatedness with other words in the segment:

$$P(w|U_k) = \lambda \frac{\rho(w|U_k)}{\sum_{z \in C_k} \rho(z|U_k)} + (1 - \lambda) \sum_{y \in C_k} \frac{e(y, w)}{\sum_{z \in C_k} e(y, z)} P(y|U_k) \qquad (10)$$

where the value of $\lambda$ ($0 \leq \lambda \leq 1$), which we call the **bias**, is a trade-off between the two components and should be set empirically. For higher values of $\lambda$, we give more weight to the word's relevance to the leading sentences compared to its relatedness with other words in the segment. Here, $C_k$ is the set of words in segment $k$, which represents the nodes in the graph. The denominators in both components are for normalization. We define $\rho(w|U_k)$ as:

$$\rho(w|U_k) = log(tf_w^{U_k} + 1).log(tf_w^k + 1) \tag{11}$$

where $tf_w^{U_k}$ and $tf_w^k$ are the number of times word $w$ appears in $U_k$ and segment $k$, respectively. A similar model has proven to be successful in measuring the relevance of a sentence to a query in query-based sentence retrieval (Allan, Wade, & Bolivar, 2003).

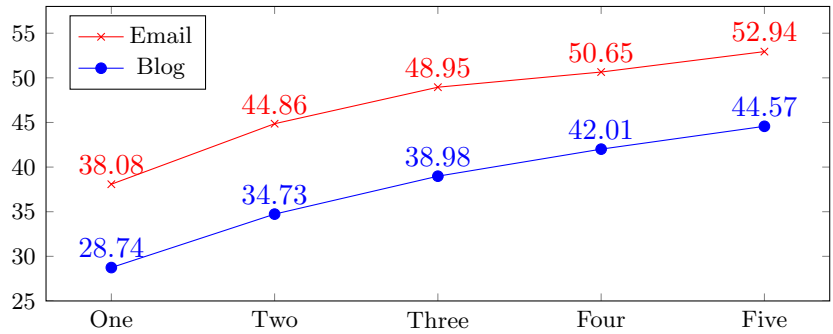

Figure 9: Percentage of words in the human-authored labels appearing in leading sentences of the topical segments.

Recall that when there are multiple topics in a conversation, a requirement for the topic labels is that labels of different topics should be discriminative (or distinguishable) (Mei et al., 2007). This implicitly indicates that a high scoring word in one segment should not have high scores in other segments of the conversation. Keeping this criterion in mind, we define the (undirected) edge weights $e(y, w)$ in equation 10 as follows:

$$e(y, w) = tf_{w,y}^k \times log \frac{K}{0.5 + tf_{w,y}^{k'}} \tag{12}$$

where $K$ denotes the number of topics (or topic segments) in the conversation, and $tf_{w,y}^k$ and $tf_{w,y}^{k'}$ are the number of times words $w$ and $y$ co-occur in a window of size $s$ in segment $k$ and in segments except $k$ in the conversation, respectively. Notice that this measure is similar in spirit to the TF.IDF metric (Salton & McGill, 1986), but it is at the co-occurrence level. The co-occurrence relationship between words captures syntactic dependencies and lexical cohesion in a text, and is also used by Mihalcea and Tarau (2004).[12]

Equation 10 above can be written in matrix notation as:

$$\pi = [\lambda Q + (1 - \lambda)R]^T \pi = A^T \pi, \tag{13}$$

where $Q$ and $R$ are square matrices such that $Q_{i,j} = \frac{\rho(j|U_k)}{\sum_{z \in C_k} \rho(z|U_k)}$ for all $i$, and $R_{i,j} = \frac{e(i,j)}{\sum_{j \in C_k} e(i,j)}$, respectively. Notice that $A$ is a stochastic matrix (i.e., all rows add up to 1), therefore, it can be treated as the transition matrix of a Markov chain. If we assume each word is a state in a Markov chain, then $A_{i,j}$ specifies the transition probability from state $i$ to state $j$ in the corresponding Markov chain. Another interpretation of $A$ can be given by a biased random walk on the graph. Imagine performing a random walk on the graph, where at every time step, with probability $\lambda$, a transition is made to the words that are relevant to the leading sentences and with probability $1 - \lambda$, a transition is made to the related words in the segment. Every transition is weighted according to the corresponding elements of $Q$ and $R$. The vector $\pi$ we are looking for is the stationary distribution of this Markov chain and is also the (normalized) eigenvector of $A$ for the eigenvalue 1. A Markov chain will have a unique stationary distribution if it is ergodic (Seneta, 1981). We can ensure the Markov chain to have this property by reserving a small probability for jumping to any other state from the current state (Page et al., 1999).[13] For larger matrices, $\pi$ can be efficiently computed by an iterative method called **power method**.

---

12. Mihalcea and Tarau (2004) use an unweighted graph for key phrase extraction. However, in our experiments, we get better results with a weighted graph.

*Incorporating Conversational Structure*

In Section 3.1, we described how the fine conversation structure in the form of a Fragment Quotation Graph (FQG) can be effectively exploited in our topic segmentation models. We hypothesize that our topic labeling model can also benefit from the FQG. In our previous work on email summarization (Carenini et al., 2008), we applied PageRank to the FQG to measure the importance of a sentence and demonstrated the benefits of using a FQG. This finding implies that an important node in the FQG is likely to cover an important aspect of the topics discussed in the conversation. Our intuition is that, to be in the topic label, a keyword should not only co-occur with other keywords, but it should also come from an important fragment in the FQG. We believe there is a mutually reinforcing relationship between the FQG and the Word Co-occurrence Graph (WCG) that should be reflected in the rankings. Our proposal is to implement this idea as a process of **co-ranking** (Zhou et al., 2007) in a heterogeneous graph, where three random walks are combined together.

Let $G = (V, E) = (V_F \cup V_W, E_F \cup E_W \cup E_{FW})$ be the heterogeneous graph of fragments and words. As shown in Figure 10, it contains three sub-graphs. First, $G_F = (V_F, E_F)$ is the unweighted directed FQG, with $V_F$ denoting the set of fragments and $E_F$ denoting the set of directed links between fragments. Second, $G_W = (V_W, E_W)$ is the weighted undirected WCG, where $V_W$ is the set of words in the segment and $E_W$ is the set of edge-weights as defined in equation 12. Third, $G_{FW} = (V_{FW}, E_{FW})$ is the weighted bipartite graph that ties $G_F$ and $G_W$ together representing the occurrence relations between the words and the fragments. Here, $V_{FW} = V_F \cup V_W$, and weighted undirected edges in $E_{FW}$ connect each fragment $v_f \in V_F$ to each word $v_w \in V_W$, with the weight representing the number of times word $v_w$ occurs in fragment $v_f$.

The co-ranking framework combines three random walks, one on $G_F$, one on $G_W$ and one on $G_{FW}$. Let $F$ and $W$ denote the transition matrices for the (intra-class) random walks in $G_F$ and $G_W$, respectively, and $\mathbf{f}$ and $\mathbf{w}$ denote their respective stationary distributions. Since, $G_{FW}$ is a bipartite graph, the (inter-class) random walk on $G_{FW}$ can be described by two transition matrices, $FW_{|V_F| \times |V_W|}$ and $WF_{|V_W| \times |V_F|}$. One intra-class step changes the probability distribution from $(\mathbf{f}, \mathbf{0})$ to $(F^T \mathbf{f}, \mathbf{0})$ or from $(\mathbf{0}, \mathbf{w})$ to $(\mathbf{0}, W^T \mathbf{w})$, while one inter-class step changes the distribution from $(\mathbf{f}, \mathbf{w})$ to $(WF^T\mathbf{w}, FW^T\mathbf{f})$ (for details see Zhou, Orshanskiy, Zha, & Giles, 2007). The coupling is regulated by a parameter $\delta$ ($0 \le \delta \le 1$) that determines the extent to which the ranking of words and fragments depend on each other. Specifically, the two update steps in the power method are:

$$f^{t+1} = (1 - \delta) \; (F^T f^t) + \delta \; WF^T(FW^TWF^T)w^t \qquad (14)$$
$$w^{t+1} = (1 - \delta) \; (W^T w^t) + \delta \; FW^T(WF^TFW^T)f^t \qquad (15)$$

---

13. For simplicity, we do not make this random jump component explicit in our equations. But, readers should keep in mind that all the transition matrices described in this article contain this component.

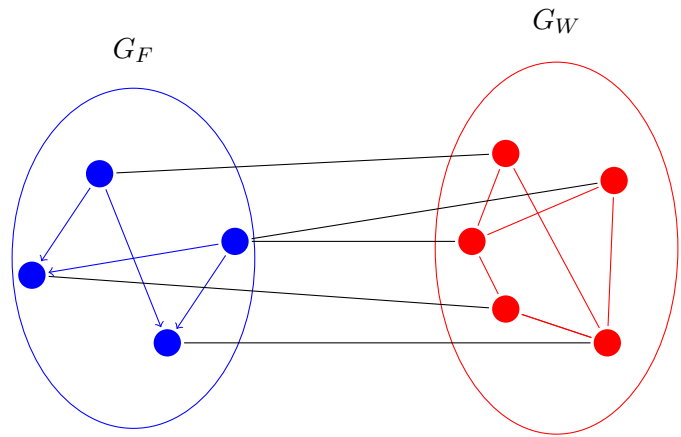

Figure 10: Three sub-graphs used for co-ranking: the fragment quotation graph $G_F$, the word co-occurrence graph $G_W$, and the bipartite graph $G_{FW}$ that ties the two together. Blue nodes represent fragments, red nodes represent words.

We described the co-ranking framework above assuming that we have a WCG and its corresponding FQG. However, recall that while the WCG is built for a topic segment, the FQG described so far (Figure 4) is based on the whole conversation. In order to construct a FQG for a topic segment in the conversation, we take only those fragments (and the edges) from the conversation-level FQG that include only the sentences of that segment. This operation has two consequences. One, some conversation-level fragments may be pruned. Two, some sentences in a conversation-level fragment may be discarded. For example, the FQG for topic (segment) ID 1 in Figure 2 includes only the fragments $a, h, i, j$, and $l$, and the edges between them. Fragment $j$, which contains three sentences in the conversation-level FQG, contains only one sentence in the FQG for topic ID 1.

### 3.2.3 Phrase Generation

Once we have a ranked list of words for describing a topical segment, we select the top $M$ keywords for constructing the keyphrases (labels) from these keywords. We take a similar approach to Mihalcea and Tarau (2004). Specifically, we mark the $M$ selected keywords in the text, and collapse the sequences of adjacent keywords into keyphrases. For example, consider the first sentence, ".. 15th anniversary of the Elder Scrolls series .." in Figure 2. If 'Elder', 'Scrolls' and 'series' are selected as keywords, since they appear adjacent in the text,

547

they are collapsed into one single keyphrase 'Elder Scrolls series'. The score of a keyphrase is then determined by taking the maximum score of its constituents (i.e., keywords).

Rather than constructing the keyphrases in the post-processing phase, as we do, an alternative approach is to first extract the candidate phrases using either n-gram sequences or a chunker in the preprocessing, and then rank those candidates (e.g., Medelyan, 2009; Hulth, 2003). However, determining the optimal value of $n$ in the n-gram sequence is an issue, and including all possible n-gram sequences for ranking excessively increases the problem size. Mei et al. (2007) also show that using a chunker leads to poor results due to the inaccuracies in the chunker, especially when it is applied to a new domain like ours.

### 3.2.4 CONVERSATION-LEVEL PHRASE RE-RANKING

So far, we have extracted phrases only from the topic segment ignoring the rest of the conversation. This method fails to find a label if some of its constituents appear outside the segment. For example, in our Blog corpus, the phrase *server security* in the human-authored label *server security and firewall* does not appear in its topical segment, but appears in the whole conversation. In fact, in our development set, about 14% and 8% words in the blog and email labels, respectively, come from the part of the conversation that is outside the topic segment. Thus, we propose to extract informative phrases from the whole conversation, re-rank those with respect to the individual topics (or segments) and combine only the relevant conversation-level phrases with the segment-level ones.

We rank the words of the whole conversation by applying the ranking models described in Section 3.2.2 and extract phrases using the same method described in Section 3.2.3. Note that when we apply our biased random walk model to the whole conversation, there is no concept of leading sentences and no distinction between the topics. Therefore, to apply to the whole conversation, we adjust our biased random walk model (Equation 10) as follows:

$$P(w) = \sum_{y \in C_k} \frac{e(y, w)}{\sum_{z \in C_k} e(y, z)} P(y) \tag{16}$$

where $e(y, w) = t f_{w,y}$, is the number of times words $w$ and $y$ co-occur in a window of size $s$ in the conversation. On the other hand, the co-ranking framework, when applied to whole conversation, combines two conversation-level graphs: the FQG, and the WCG.

To re-rank the phrases extracted from the whole conversation with respect to a particular topic in the conversation, we reuse the score of the words in that topic segment (given by the ranking models in Section 3.2.2). As before, the score of a (conversation-level) phrase is determined by taking the maximum (segment-level) score of its constituents (words). If a word does not occur in the topic segment, its score is assumed to be 0.

### 3.2.5 REDUNDANCY CHECKING

Once we have the ranked list of labels (keyphrases), the last step is to produce the final $k$ labels as output. When selecting multiple labels for a topic, we expect the new labels to be diverse without redundant information to achieve broad coverage of the topic. We use the Maximum Marginal Relevance (MMR) (Carbonell & Goldstein, 1998) criterion to select the labels that are relevant, but not redundant. Specifically, we select the labels one by one, by maximizing the following MMR criterion each time:

$$\hat{l} = argmax_{l \in W-S}[\rho \ Score(l) - (1 - \rho) \ max_{\hat{l} \in S} \ Sim(\hat{l}, l)] \tag{17}$$

where $W$ is the set of all labels and $S$ is the set of labels already selected as output. We define the similarity between two labels $\hat{l}$ and $l$ as: $Sim(\hat{l}, l) = n_o/n_l$, where $n_o$ is the number of overlapping (modulo stemming) words between $\hat{l}$ and $l$, and $n_l$ is the number of words in $l$. The parameter $\rho$ $(0 \leq \rho \leq 1)$ quantifies the amount of redundancy allowed.

## 4. Corpora and Metrics

Due to the lack of publicly available corpora of asynchronous conversations annotated with topics, we have developed the first corpora annotated with topic information.

### 4.1 Data Collection

For email, we selected our publicly available BC3 email corpus (Ulrich, Murray, & Carenini, 2008) which contains 40 email conversations from the World Wide Web Consortium (W3C) mailing list[14]. The BC3 corpus, previously annotated with sentence-level speech acts, subjectivity, extractive and abstractive summaries, is one of a growing number of corpora being used for email research (Carenini et al., 2011). This has an average of 5 emails per conversation and a total of 1024 sentences after excluding the quoted sentences. Each conversation also provides the thread structure based on reply-to relations between emails.

For blog, we manually selected 20 conversations of various lengths, all short enough to still be feasible for humans to annotate, from the popular technology-related news website Slashdot[15]. Slashdot was selected because it provides reply-to links between comments, allowing accurate thread reconstruction, and since the comments are moderated by the users of the site, they are expected to have a decent standard. A conversation in Slashdot begins with an article (i.e., a short synopsis paragraph possibly with a link to the original story), and is followed by a lengthy discussion section containing multiple threads of comments and single comments. This is unlike an email conversation which contains a single thread of emails. The main article is assumed to be the root in the conversation tree (based on reply-to), while the threads and the single comments form the sub-trees in the tree. In our blog corpus, we have a total of $4,411$ sentences. The total number of comments per blog conversation varies from 30 to 101 with an average of 60.3, the number of threads per conversation varies from 3 to 16 with an average of 8.35 and the number of single comments varies from 5 to 50 with an average of 20.25.

### 4.2 Topic Annotation

Topic segmentation and labeling in general is a nontrivial and subjective task even for humans, particularly when the text is unedited and less organized (Purver, 2011). The conversation phenomenon called **'Schism'** makes it even more challenging for conversations. During a schism, a new conversation takes birth from an existing one, not necessarily because of a topic shift but because some participants refocus their attention onto each other, and away from whoever held the floor in the parent conversation (Sacks, Schegloff, & Jefferson, 1974). In the example email conversation shown in Figure 1, a schism takes place when the participants discuss the topic 'responding to I18N'. Not all our annotators agree on the fact that the topic 'responding to I18N' swerves from the topic 'TAG document'.

---

14. http://research.microsoft.com/en-us/um/people/nickcr/w3c-summary.html
15. http://slashdot.org/

To properly design an effective annotation manual and procedure, we performed a two-phase pilot study before carrying out the actual annotation. Our initial annotation manual was inspired by the AMI annotation manual used for topic segmentation of ICSI meeting transcripts[16]. For the pilot study, we selected two blog conversations from Slashdot and five email conversations from the W3C corpus. Note that these conversations were not picked from our corpora. Later in our experiments we use these conversations as our **development set** for tuning different parameters of the computational models. In the first phase of the pilot study five computer science graduate students volunteered to do the annotation, generating five different annotations for each conversation. We then revised our annotation manual based on their feedback and a detailed analysis of possible sources of disagreement. In the second phase, we tested our procedure with a university postdoc doing the annotation.

We prepared two different annotation manuals – one for email and one for blog. We chose to do so for mainly two reasons. (i) As discussed earlier, our email and blog conversations are structurally different and have their own specific characteristics. (ii) The email corpus already had some annotations (e.g., abstract summaries) that we could reuse for topic annotation, whereas our blog corpus is brand new without any existing annotation.

For the actual annotation we recruited and paid three cognitive science fourth year under-graduates, who are native speakers of English and also Slashdot bloggers. On average, they took about 7 and 28.5 hours to annotate the 40 email and 20 blog conversations, respectively. In all, we have three different annotations for each conversation in our corpora. For blog conversations, the task of finding topics was carried out in four steps:

1. The annotators read the conversation (i.e., article, threads of comments and single comments) and wrote a short summary ($\leq 3$ sentences) only for the threads.

2. They provided short high-level descriptions for the topics discussed in the conversation (e.g., 'Game contents or size', 'Bugs or faults'). These descriptions serve as reference **topic labels** in our work. The target number of topics and their labels were not given in advance and they were instructed to find as many or as few topics as needed to convey the overall content of the conversation.

3. They assigned the most appropriate topic to each sentence. However, if a sentence covered more than one topic, they labeled it with all the relevant topics according to their order of relevance. They used the predefined topic 'OFF-TOPIC' if the sentence did not fit into any topic. Wherever appropriate they also used two other predefined topics: 'INTRO' (e.g., 'hi X') and 'END' (e.g., 'Best, X').

4. The annotators authored a single high-level 250 words summary of the whole conversation. This step was intended to help them remember anything they may have forgotten and to revise the annotations in the previous three steps.

---

16. http://mmm.idiap.ch/private/ami/annotation/TopicSegmentationGuidelinesNonScenario.pdf

For each email conversation in BC3, we already had three human-authored summaries. So, along with the actual conversations, we provided the annotators with such summaries to give them a brief overview of the discussion. After reading a conversation and the associated summaries, they performed tasks 2 and 3 as in the procedure they follow for annotating blogs. The annotators carried out the tasks on paper. We created the hierarchical thread view of the conversation based on the reply-to relations between the comments (or emails) using indentations and printed each participant's information in a different color as in Gmail.

In the email corpus, the three annotators found 100, 77 and 92 topics respectively (269 in total), and in the blog corpus, they found 251, 119 and 192 topics respectively (562 in total). Table 3 shows some basic statistics computed on the three annotations of the conversations.[17] On average, we have 26.3 sentences and 2.5 topics per email conversation, and 220.55 sentences and 10.77 topics per blog conversation. On average, a topic in email conversations contains 12.6 sentences, and a topic in blog conversations contains 27.16 sentences. The average number of topics active at a time are 1.4 and 5.81 for email and blog conversations, respectively. The average **entropy** which corresponds to the granularity of an annotation (as described in the next Section) is 0.94 for email conversations and 2.62 for blog conversations. These statistics (i.e., the number of topics and the topic density) indicate that there is a substantial amount of segmentation (and labeling) to do.

|  | Mean | | Max | | Min | |
|---|---|---|---|---|---|---|
|  | **Email** | **Blog** | **Email** | **Blog** | **Email** | **Blog** |
| Number of sentences | 26.3 | 220.55 | 55 | 430 | 13 | 105 |
| Number of topics | 2.5 | 10.77 | 7 | 23 | 1 | 5 |
| Average topic length | 12.6 | 27.16 | 35 | 61.17 | 3 | 11.67 |
| Average topic density | 1.4 | 5.81 | 3.1 | 10.12 | 1 | 2.75 |
| Entropy | 0.94 | 2.62 | 2.7 | 3.42 | 0 | 1.58 |

Table 3: Statistics on three human annotations per conversation.

## 4.3 Evaluation (and Agreement) Metrics

In this section we describe the metrics used to compare different annotations. These metrics measure both how much our annotators agree with each other, and how well our models and various baselines perform. For a given conversation, different annotations can have different numbers of topics, different topic assignments of the sentences (i.e., the clustering) and different topic labels. Below we describe the metrics used to measure the segmentation performance followed by the metrics used to measure the labeling performance.

### 4.3.1 Metrics for Topic Segmentation

As different annotations can group sentences in different numbers of clusters, agreement metrics widely used in supervised classification, such as the $\kappa$ statistic and $F_1$ score, are not applicable. Again, our problem of topic segmentation in asynchronous conversation is not sequential in nature. Therefore, the standard metrics widely used in sequential topic segmentation in monolog and synchronous dialog, such as the $P_k$ (Beeferman, Berger, & Lafferty, 1999) and $WindowDiff(WD)$ (Pevzner & Hearst, 2002), are also not applicable. Rather, the **one-to-one** and **local agreement** metrics described by Elsner and Charniak (2010) are more appropriate for our segmentation task.

---

17. We got 100% agreement on the two predefined topics 'INTRO' and 'END'. Therefore, in all our computations we excluded the sentences marked as either 'INTRO' or 'END'.

The one-to-one metric measures global agreement between two annotations by pairing up topical segments from the two annotations in a way (i.e., by computing the optimal max-weight bipartite matching) that maximizes the total overlap, and then reports the percentage of overlap. The local agreement metric $loc_k$ measures agreement within a context of $k$ sentences. To compute the $loc_3$ score for the m-th sentence in the two annotations, we consider the previous 3 sentences: m-1, m-2 and m-3, and mark them as either 'same' or 'different' depending on their topic assignment. The $loc_3$ score between two annotations is the mean agreement on these 'same' or 'different' judgments, averaged over all sentences. See Appendix A for a detailed description of these metrics with concrete examples.

We report the annotators' agreement found in one-to-one and $loc_3$ metrics in Table 4. For each human annotation, we measure its agreement with the two other human annotations separately, and report the mean agreements. For email, we get high agreement in both metrics, though the local agreement (average of 83%) is a little higher than the global one (average of 80%). For blog, the annotators have high agreement in $loc_3$ (average of 80%), but they disagree more in one-to-one (average of 54%). A low one-to-one agreement in blog is quite acceptable since blog conversations are much longer and less focused than email conversations (see Table 3). By analyzing the two corpora we also noticed that in blogs, people are more informal and often make implicit jokes (see Figure 2). As a result, the segmentation task in blogs is more challenging for humans as well as for our models. Note that in a similar annotation task for chat disentanglement, Elsner and Charniak (2010) report an average one-to-one score of 53%. Since the one-to-one score for naive baselines (see Section 5.1) is much lower than the human agreement, this metric differentiates human-like performance from baseline. Therefore, computing one-to-one correlation with the human annotations is a legitimate evaluation for our models.

|  | Mean | | Max | | Min | |
|---|---|---|---|---|---|---|
|  | **Email** | **Blog** | **Email** | **Blog** | **Email** | **Blog** |
| one-to-one | 80.4 | 54.2 | 100.0 | 84.1 | 31.3 | 25.3 |
| $loc_3$ | 83.2 | 80.1 | 100.0 | 94.0 | 43.7 | 63.3 |

Table 4: Annotator agreement in one-to-one and $loc_3$ on the two corpora.

When we analyze the source of disagreement in the annotation, we find that by far the most frequent reason is the same as the one observed by Elsner and Charniak (2010) for the chat disentanglement task; namely, some annotators are more specific (i.e., fine) than others (i.e., coarse). To determine the level of specificity in an annotation, similarly to Elsner and Charniak, we use the information-theoretic concept of **entropy**. If we consider the topic of a randomly picked sentence in a conversation as a random variable $X$, its entropy $H(X)$ measures the level of details in an annotation. For topics $k$ each having length $n_k$ in a conversation of length $N$, we compute $H(X)$ as follows:

$$H(X) = -\sum_{k=1}^{K} \frac{n_k}{N} log_2 \frac{n_k}{N} \qquad (18)$$

where $K$ is the total number of topics (or topical segments) in the conversation. The entropy gets higher as the number of topics increases and the topics are evenly distributed in a conversation. In our corpora, it varies from 0 to 2.7 in email conversations and from 1.58 to 3.42 in blog conversations (Table 3). These variations demonstrate the differences in specificity for different annotators, but do not determine their agreement on the general structure. To quantify this, we use the **many-to-one** metric proposed by Elsner and Charniak (2010). It maps each of the source clusters to the single target cluster with which it gets the highest overlap, then computes the total percentage of overlap. This metric is asymmetrical, and not to be used for performance evaluation.[18] However, it provides some insights about the annotation specificity. For example, if one splits a cluster of another annotator into multiple sub-clusters then, the many-to-one score from fine to coarse annotation is 100%. In our corpora, by mapping from fine (high-entropy) to coarse (low-entropy) annotation we get high many-to-one score, with an average of 95% in email conversations and an average of 72% in blog conversations (Table 5). This suggests that the finer annotations have mostly the same scopic boundaries as the coarser ones.

|  | Mean | | Max | | Min | |
|---|---|---|---|---|---|---|
|  | **Email** | **Blog** | **Email** | **Blog** | **Email** | **Blog** |
| many-to-one | 94.9 | 72.3 | 100 | 98.2 | 61.1 | 51.4 |

Table 5: Annotator agreement in many-to-one on the two corpora.

### 4.3.2 Metrics for Topic Labeling

Recall that we extract keyphrases from the text as topic labels. Traditionally keyphrase extraction is evaluated using precision, recall and F-measure based on exact matches between the extracted keyphrases and the human-assigned keyphrases (e.g., Mihalcea and Tarau, 2004; Medelyan et al., 2009). However, it has been noted that this approach based on exact matches underestimates the performance (Turney, 2000). For example, when compared with the reference keyphrase 'Game contents or size', a credible candidate keyphrase 'Game contents' gets evaluated as wrong in this metric. Therefore, recent studies (Zesch & Gurevych, 2009; Kim, Baldwin, & Kan, 2010a) suggest to use the $n$-gram-based metrics that account for near-misses, similar to the ones used in text summarization, e.g., ROUGE (Lin, 2004), and machine translation, e.g., BLEU (Papineni, Roukos, Ward, & Zhu, 2002).

Kim et al. (2010a) evaluated the utility of different $n$-gram-based metrics for keyphrase extraction and showed that the metric which we call **mutual-overlap (m-o)**, correlates most with human judgments.[19] Therefore, one of the metrics we use for evaluating our topic labeling models is m-o. Given a reference keyphrase $p_r$ of length (in words) $n_r$, a candidate keyphrase $p_c$ of length $n_c$, and $n_o$ being the number of overlapping (modulo stemming) words between $p_r$ and $p_c$, mutual-overlap is formally defined as:

$$mutual-overlap(p_r, p_c) = \frac{n_o}{max(n_r, n_c)} \qquad (19)$$

---

18. One can easily optimize it by assigning a different topic to each of the source sentences.

19. Kim et al. (2010a) call this metric **R-precision (R-p)**, which is different from the actual definition of R-p for keyphrase evaluation given by Zesch and Gurevych (2009). Originally, R-p is the precision measured when the number of candidate keyphrases equals the number of gold keyphrases.

This metric gives full credit to exact matches and morphological variants, and partial credit to two cases of overlapping phrases: (i) when the candidate keyphrase includes the reference keyphrase, and (ii) when the candidate keyphrase is a part of the reference keyphrase. Notice that m-o as defined above evaluates a single candidate keyphrase against a reference keyphrase. In our setting, we have a single reference keyphrase (i.e., topic label) for each topical cluster, but as mentioned before, we may want our models to extract the top $k$ keyphrases. Therefore, we modify m-o to evaluate a set of $k$ candidate keyphrases $P_c$ against a reference keyphrase $p_r$ as follows, calling it **weighted-mutual-overlap (w-m-o)**:

$$weighted-mutual-overlap(p_r, P_c) = \sum_{i=1}^{k} \frac{n_o}{max(n_r, n_c^i)} S(p_c^i) \qquad (20)$$

where $S(p_c^i)$ is the normalized score (i.e., $S(p_c^i)$ satisfies $0 \leq S(p_c^i) \leq 1$ and $\sum_{i=1}^{k} S(p_c^i) = 1$) of the i-th candidate phrase $p_c^i \in P_c$. For $k = 1$, this metric is equivalent to m-o, and for higher values of $k$, it takes the sum of $k$ m-o scores, each weighted by its normalized score.

The w-m-o metric described above only considers word overlap and ignores other semantic relations (e.g., synonymy, hypernymy) between words. However, annotators when writing the topic descriptions, may use words that are not directly from the conversation, but are semantically related. For example, given a reference keyphrase 'meeting agenda', its lexical semantic variants like 'meeting schedule' or 'meeting plan' should be treated as correct. Therefore, we also consider a generalization of w-m-o that incorporates lexical semantics. We define **weighted-semantic-mutual-overlap (w-s-m-o)** as follows:

$$weighted-semantic-mutual-overlap(p_r, P_c) = \sum_{i=1}^{k} \frac{\sum_{t_r \in p_r} \sum_{t_c \in p_c^i} \sigma(t_r, t_c)}{max(n_r, n_c^i)} S(p_c^i) \qquad (21)$$

where $\sigma(t_r, t_c)$ is the semantic similarity between the nouns $t_r$ and $t_c$. The value of $\sigma(t_r, t_c)$ is between 0 and 1, where 1 denotes notably high similarity and 0 denotes little-to-none. Notice that, since this metric considers semantic similarity between all possible pairs of nouns, the value of this measure can be greater than 100% (when presented in percentage). We use the metrics (e.g., lin_similarity, wup_similarity) provided in the WordNet::Similarity package (Pedersen, Patwardhan, & Michelizzi, 2004) for computing WordNet-based similarity, and always choose the most frequent sense for a noun. The results we get are similar across the similarity metrics. For brevity, we just mention the lin_similarity in this article.

### 4.3.3 Metrics for End-to-End Evaluation

Just like the human annotators, our end-to-end system takes an asynchronous conversation as input, finds the topical segments in the conversation, and then assigns short descriptions (topic labels) to each of the topical segments. It would be fairly easy to compute agreement on topic labels based on mutual overlaps, if the number of topics and topical segments were fixed across the annotations of a given conversation. However, since different annotators (system or human) can identify a different number of topics and different clustering of sentences, measuring annotator (model or human) agreement on the topic labels is not a trivial task. To solve this, we first map the clusters of one annotation (say $A_1$) to the clusters of another (say $A_2$) by the optimal one-to-one mapping described in the previous section. After that, we compute the w-m-o and w-s-m-o scores on the labels of the mapped (or paired) clusters. Formally, if $l_i^1$ is the label of cluster $c_i^1$ in $A_1$ that is mapped to the cluster $c_j^2$ with label $l_j^2$ in $A_2$, we compute w-m-o$(l_i^1, l_j^2)$ and w-s-m-o$(l_i^1, l_j^2)$.

Table 6 reports the human agreement for w-m-o and w-s-m-o on the two corpora. Similar to segmentation, we get higher agreement on labeling for both metrics on email. Plausibly, the reasons remain the same; the length and the characteristics (e.g., informal, less focused) of blog conversations make the annotators disagree more. However, note that these measures are computed based on one-to-one mappings of the clusters and may not reflect the same agreement one would get if the annotators were asked to label the same segments.

|           | Mean      |        | Max       |        | Min       |        |
|-----------|-----------|--------|-----------|--------|-----------|--------|
|           | **Email** | **Blog** | **Email** | **Blog** | **Email** | **Blog** |
| w-m-o     | 36.8      | 19.9   | 100.0     | 54.2   | 0.0       | 0.0    |
| w-s-m-o   | 42.5      | 28.2   | 107.3     | 60.8   | 0.0       | 5.2    |

Table 6: Annotator agreement in w-m-o and w-s-m-o on the two corpora.

# 5. Experiments

In this section we present our experimental results. First, we show the performance of the segmentation models. Then we show the performance of the topic labeling models based on manual segmentation. Finally, we present the performance of the end-to-end system.

## 5.1 Topic Segmentation Evaluation

In this section we present the experimental setup and results for the segmentation task.

### 5.1.1 Experimental Setup for Segmentation

We ran six different topic segmentation models on our corpora presented in Section 4. Our first model is the graph-based unsupervised segmentation model presented by Malioutov and Barzilay (2006). Since the sequentiality constraint of topic segmentation in monolog and synchronous dialog does not hold in asynchronous conversation, we implement this model without this constraint. Specifically, this model (call it **M&B**) constructs a weighted undirected graph $G(V, E)$, where the nodes $V$ represent the sentences and the edge weights $w(x, y)$ represent the cosine similarity (Equation 5) between sentences $x$ and $y$. It then finds the topical segments by optimizing the normalized cut criterion (Equation 6). Thus, M&B considers the conversation globally, but models only lexical similarity.

The other five models are LDA, LDA+FQG, LCSeg, LCSeg+FQG and the Supervised model (SUP) as described in Section 3. The tunable parameters of the different models were set based on their performance on our developement set. The hyperparameters $\alpha$ and $\beta$ in LDA were set to their default values ($\alpha$=50/$K$, $\beta$=0.01) as suggested by Steyvers and Griffiths (2007).[20] The regularization strength $\lambda$ in LDA+FQG was set to 20. The parameters of LCSeg were set to their default values since this setting delivers the best performance on the development set. For a fair comparison, we set the same number of topics per conversation in all of the models. If at least two of the three annotators agree on the topic number, we set that number, otherwise we set the floor value of the average topic number. The mean statistics of the six model annotations are shown in Table 7. Comparing with the statistics of the human annotations in Table 3, we can notice that these numbers are within the bounds of the human annotations.[21]

|  |  | M&B | LDA | LDA+FQG | LCSeg | LCSeg+FQG | SUP |
|---|---|---|---|---|---|---|---|
| **Email** | Topic number | 2.41 | 2.10 | 1.90 | 2.41 | 2.41 | 2.41 |
|  | Topic length | 12.41 | 13.3 | 15.50 | 12.41 | 12.41 | 12.41 |
|  | Topic density | 1.90 | 1.83 | 1.60 | 1.01 | 1.39 | 1.42 |
|  | Entropy | 0.99 | 0.98 | 0.75 | 0.81 | 0.93 | 0.98 |
| **Blog** | Topic number | 10.65 | 10.65 | 10.65 | 10.65 | 10.65 | 10.65 |
|  | Topic length | 20.32 | 20.32 | 20.32 | 20.32 | 20.32 | 20.32 |
|  | Topic density | 7.38 | 9.39 | 8.32 | 1.00 | 5.21 | 5.30 |
|  | Entropy | 2.54 | 3.33 | 2.37 | 2.85 | 2.81 | 2.85 |

Table 7: Mean statistics of different model's annotation

We also evaluate the following baselines, which any useful model should outperform.

- **All different** Each sentence in the conversation constitutes a separate topic.

- **All same** The whole conversation constitutes a single topic.

- **Speaker** The sentences from each participant constitute a separate topic.

- **Blocks of** $k$ **(= 5, 10, 15, 20, 25, 30):** Each consecutive group of $k$ sentences in the temporal order of the conversation constitutes a separate topic.

### 5.1.2 Results for Segmentation

Table 8 presents the human agreement and the agreement of the models with the human annotators on our corpora. For each model annotation, we measure its agreement with the three human annotations separately using the metrics described in Section 4.3.1, and report the mean agreements. In the table, we also show the performance of the two best baselines– *the Speaker* and *the Blocks of* $k$.

---

20. The performance of LDA does not seem to be sensitive to the values of $\alpha$ and $\beta$.

21. Although the topic numbers per conversation are fixed for different models, LDA and LDA+FQG may find less number of topics (see Equation 3 and 4).

|  |  | Baselines | | Models | | | | | | Human |
|---|---|---|---|---|---|---|---|---|---|---|
|  |  | Speaker | Blocks of $k$ | M&B | LDA | LDA+ FQG | LCSeg | LCSeg +FQG | SUP |  |
| **Email** | Mean 1-to-1 | 51.8 | 38.3 | 62.8 | 57.3 | 61.5 | 62.2 | **69.3** | **72.3** | 80.4 |
|  | Max 1-to-1 | 94.3 | 77.1 | 100.0 | 100.0 | 100.0 | 100.0 | 100.0 | 100.0 | 100.0 |
|  | Min 1-to-1 | 23.4 | 14.6 | 36.3 | 24.3 | 24.0 | 33.1 | 38.0 | 42.4 | 31.3 |
|  | Mean $loc_3$ | 64.1 | 57.4 | 62.4 | 54.1 | 60.6 | 72.0 | **72.7** | **75.8** | 83.2 |
|  | Max $loc_3$ | 97.0 | 73.1 | 100.0 | 100.0 | 100.0 | 100.0 | 100.0 | 100.0 | 100.0 |
|  | Min $loc_3$ | 27.4 | 42.6 | 36.3 | 38.1 | 38.4 | 40.7 | 40.6 | 40.4 | 43.7 |
| **Blog** | Mean 1-to-1 | 33.5 | 32.0 | 30.0 | 25.2 | 28.0 | 36.6 | **46.7** | **48.5** | 54.2 |
|  | Max 1-to-1 | 61.1 | 46.0 | 45.3 | 42.1 | 56.3 | 53.6 | 67.4 | 66.1 | 84.1 |
|  | Min 1-to-1 | 13.0 | 15.6 | 18.2 | 15.3 | 16.1 | 23.7 | 26.6 | 28.4 | 25.3 |
|  | Mean $loc_3$ | 67.0 | 52.8 | 54.1 | 53.0 | 55.4 | 56.5 | **75.1** | **77.2** | 80.1 |
|  | Max $loc_3$ | 87.1 | 68.4 | 64.3 | 65.6 | 67.1 | 76.0 | 89.0 | 96.4 | 94.0 |
|  | Min $loc_3$ | 53.4 | 42.3 | 45.1 | 38.6 | 46.3 | 43.1 | 56.7 | 63.2 | 63.3 |

Table 8: Topic segmentation performance of the two best Baselines, Human and Models. In the Blocks of $k$ column, $k = 5$ for email and $k = 20$ for blog.

Most of the baselines perform rather poorly. *All different* is the worst baseline of all with mean one-to-one scores of only 0.05 and 0.10, and mean $loc_3$ scores of only 0.47 and 0.25 in the blog and email corpus, respectively. *Blocks of 5* is one of the best baselines in email, but it performs poorly in blog with mean one-to-one of 0.19 and mean $loc_3$ of 0.54. On the contrary, *Blocks of 20* is one of the best baselines in blog, but performs poorly in email. This is intuitive since the average number of topics and topic length in blog conversations (10.77 and 27.16) are much higher than those of email (2.5 and 12.6). *All same* is optimal for conversations containing only one topic, but its performance rapidly degrades as the number of topics increases. It has mean one-to-one scores of 0.29 and 0.28 and mean $loc_3$ scores of 0.53 and 0.54 in the blog and email corpora, respectively. *Speaker* is the strongest baseline in both domains.[22] In several cases it beats some of the under-performing models.

In the email corpus, in one-to-one, generally the models agree with the annotators more than the baselines do, but less than the annotators agree with each other. We observe a similar trend in the local metric $loc_3$, however on this metric, some models fail to beat the best baselines. Notice that human agreement for some of the annotations is quite low (see the Min scores), even lower than the mean agreement of the baselines. As explained before, this is due to the fact that some human annotations are much more fine-grained than others.

In the blog corpus, the agreement on the global metric (one-to-one) is much lower than that on the email corpus. The reasons were already explained in Section 4.3.1. We notice a similar trend in both metrics– some under-performing models fail to beat the baselines, while others perform better than the baselines, but worse than the human annotators.

The comparison among the models reveals a general pattern. The probabilistic generative models LDA and LDA+FQG perform disappointingly on both corpora. A likely explanation is that the independence assumption made by these models when computing the distribution over topics for a sentence from the distributions of its words causes nearby sentences (i.e., local context) to be excessively distributed over topics. Another reason could be the limited amount of data available for training. In our corpora, the average number of sentences per blog conversation is 220.55 and per email conversation is 26.3, which might not be sufficient for the LDA models (Murphy, 2012). If we compare the performance of LDA+FQG with the performance of LDA, we get a significant improvement with LDA+FQG in both metrics on both corpora (p<0.01). The regularization with the FQG prevents the local context from being excessively distributed over topics.

---

22. There are many anonymous authors in our blog corpus. We treated each of them as a separate author.

The unsupervised graph-based model M&B performs better than the LDA models in most cases (i.e., except $loc_3$ in blog) ($p < 0.001$). However, its performance is still far below the performance of the top performing models like LCSeg+FQG and the supervised model. The reason is that even though, by constructing a complete graph, this method considers the conversation globally, it only models the lexical similarity and disregards other important features of asynchronous conversation like the fine conversation structure and the speaker.

Comparison of LCSeg with LDAs and M&B reveals that LCSeg in general is a better model. LCSeg outperforms LDA by a wide margin in one-to-one on two datasets and in $loc_3$ on email ($p < 0.001$). The difference between LCSeg and LDA in $loc_3$ on blog is also significant with $p < 0.01$. LCSeg also outperforms M&B in most cases ($p < 0.01$) except in one-to-one on email. Since LCSeg is a sequential model it extracts the topics keeping the context intact. This helps it to achieve high $loc_3$ agreement for shorter conversations like email conversations. But, for longer conversations like blog conversations, it overdoes this (i.e., extracts larger chunks of sentences as a topic segment) and gets low $loc_3$ agreement. This is unsurprising if we look at its topic density in Table 7 on the two datasets– the density is very low in the blog corpus compared to annotators and other well performing models. Another reason of its superior performance over LDAs and M&B could be its term weighting scheme. Unlike LDAs and M&B, which consider only repetition, LCSeg also considers how tightly the repetition happens. However, there is still a large gap in performance between LCSeg and other top performing models (LCSeg+FQG, the supervised). As explained earlier, topics in an asynchronous conversation may not change sequentially in the temporal order of the sentences. If topics are interleaved then LCSeg fails to identify them correctly. Furthermore, LCSeg does not consider other important features beyond the lexical cohesion.

When we incorporate FQG into LCSeg, we get a significant improvement in one-to-one on both corpora and in $loc_3$ on blog (p<0.0001). Even though the improvement in $loc_3$ on email is not significant, the agreement is quite high compared to other unsupervised models. Overall, LCSeg+FQG is the best unsupervised model. This supports our claim that sentences connected by reply-to relations in the FQG usually refer to the same topic.

Finally, when we combine all the features into our graph-based supervised model (SUP in Table 8), we get a significant improvement over LCSeg+FQG in both metrics across both domains (p<0.01). The agreements achieved by the supervised model are also much closer to that of human annotators. Beside the features, this improvement might also be due to the fact that, by constructing a complete graph, this model considers relations between all possible sentence pairs in a conversation, which we believe is a key requirement for topic segmentation in asynchronous conversations.

## 5.2 Topic Labeling Evaluation

In this section we present the experimental evaluation of the topic labeling models when the models are provided with manual (or gold) segmentation. This allows us to judge their performance independently of the topic segmentation task.

### 5.2.1 Experimental Setup for Topic Labeling

As mentioned in Section 4, in the email corpus, the three annotators found 100, 77 and 92 topics (or topical segments) respectively (269 in total), and in the blog corpus, they found 251, 119 and 192 topics (562 in total). The annotators wrote a short high-level description for each topic. These descriptions serve as reference topic labels in our evaluation.[23] The goal of the topic labeling models is to automatically generate such informative descriptions for each topical segment. We compare our approach with two baselines. The first baseline **FreqBL** ranks the words according to their frequencies. The second baseline **LeadBL**, expressed by equation 11, ranks the words based on their relevance only to the leading sentences in a topical segment.

We also compare our model with two state-of-the-art keyphrase extraction methods. The first one is the unsupervised general TextRank model proposed by Mihalcea and Tarau (2004) (call it **M&T**) that does not incorporate any conversation specific information. The second one is the supervised model **Maui** proposed by Medelyan et al. (2009). Briefly, Maui first extracts all n-grams up to a maximum length of 3 as candidate keyphrases. Then a bagged decision tree classifier filters the candidates using nine different features. Due to the lack of labeled training data in asynchronous conversations, we train Maui on the human-annotated dataset released as part of the SemEval-2010 task 5 on automatic keyphrase extraction from scientific articles (Kim, Medelyan, Kan, & Baldwin, 2010b). This dataset contains 244 scientific papers from the ACM digital library, each comes with a set of author-assigned and reader-assigned keyphrases. The total number of keyphrases assigned to the 244 articles by both the authors and the readers is 3705.

We experimented with two different versions of our biased random walk model that incorporates informative clues from the leading sentences. One, **BiasRW**, does not include any conversation-level phrase (Section 3.2.4), and the other one **BiasRW+**, does. The parameter $U_k$, the set of leading sentences, was empirically set to the first two sentences and the bias parameter $\lambda$ was set to 0.85 based on our development set.

We experimented with four different versions of the co-ranking framework depending on what type of random walk is performed on the word co-occurrence graph (WCG) and whether the model includes any conversation-level phrases. Let **CorGen** denote the co-ranking model with a general random walk on WCG, and **CorBias** denote the co-ranking model with a biased random walk on WCG. These two models do not include any conversation-level phrase while **CorGen+** and **CorBias+** do. The coupling strength $\delta$ and the co-occurrence window size $s$ were empirically set to 0.4 and 2, respectively, based on the development set. The dumping factor was set to its default value 0.85.

Note that all the models (except Maui) and the baselines follow the same preprocessing and post-processing (i.e., phrase generation and redundancy checking) steps. The value of $M$ in phrase generation was set to 25% of the total number of words in the cluster, and $\rho$ in redundancy checking was set to 0.35 based on the development set.

---

23. Notice that in our setting, for each topic segment we have only one reference label to compare with. Therefore, we do not show the human agreement on the topic labeling task in Table 9 and 10.

### 5.2.2 RESULTS FOR TOPIC LABELING

We evaluate the performance of different models using the metrics described in Section 4.3.2. Table 9 and 10, respectively, show the mean weighted-mutual-overlap (w-m-o) and weighted-semantic-mutual-overlap (w-s-m-o) scores in percentage of different models for different values of $k$ (i.e., number of output labels) on the two corpora.

Both the baselines have proved to be strong, beating the existing models in almost every case. This tells us that the frequency of the words in the topic segment and their occurrence in the leading sentences carry important information for topic labeling. Generally speaking, LeadBL is a better baseline for email, while for blog FreqBL is better than LeadBL.

The supervised model Maui is the worst performer in both metrics on the two corpora. Its performance is also consistently low across the corpora for any particular value of $k$. A possible explanation is that Maui was trained on a domain (scientific articles), which is rather different from asynchronous conversations. Another reason may be that Maui does not consider any conversational features.

The general random walk model M&T also delivers poor performance on our corpora, failing to beat the baselines in both measures. This indicates that the random walk model based on only co-occurrence relations between the words is not sufficient for finding topic labels in asynchronous conversations. It needs to consider conversation specific information.

By incorporating the clues from the leading sentences, our biased random walk model BiasRW improves the performance significantly over the baselines in both metrics for all the values of $k$ on the two corpora ($p<0.05$). This demonstrates the usefulness of considering the leading sentences as an information source for topic labeling in asynchronous conversation.

| | | k=1 | | k=2 | | k=3 | | k=4 | | k=5 | |
|---|---|---|---|---|---|---|---|---|---|---|---|
| | | Email | Blog | Email | Blog | Email | Blog | Email | Blog | Email | Blog |
| **Baselines** | FreqBL | 22.86 | 19.05 | 17.47 | 16.17 | 14.96 | 13.83 | 13.17 | 13.45 | 12.06 | 12.59 |
| | LeadBL | 22.41 | 18.17 | 18.94 | 15.95 | 15.92 | 13.75 | 14.36 | 12.61 | 13.76 | 11.93 |
| **Models** | M&T | 15.87 | 18.23 | 12.68 | 14.31 | 10.33 | 12.15 | 9.63 | 11.38 | 9.07 | 11.03 |
| | Maui | 10.48 | 10.03 | 9.86 | 9.56 | 9.03 | 9.23 | 8.71 | 8.90 | 8.50 | 8.53 |
| | BiasRW | 24.77 | 20.83 | 19.78 | 17.28 | 17.38 | 15.06 | 16.24 | 14.53 | 15.80 | 14.26 |
| | BiasRW+ | 24.91 | 23.65 | **20.36** | 19.69 | 18.09 | 17.76 | 16.20 | 16.78 | 15.78 | 15.86 |
| | CorGen | 17.60 | 20.76 | 15.32 | 17.64 | 15.14 | 15.78 | 14.23 | 15.03 | 14.08 | 14.75 |
| | CorGen+ | 18.32 | 22.44 | 15.86 | 19.65 | 15.46 | 18.01 | 14.89 | 16.90 | 14.45 | 16.13 |
| | CorBias | 24.84 | 20.96 | 19.88 | 17.73 | 17.61 | 16.22 | 16.99 | 15.64 | 16.81 | 15.38 |
| | CorBias+ | **25.13** | 23.83 | 20.20 | **19.97** | **18.21** | **18.33** | **17.15** | **17.28** | **16.90** | **16.55** |

Table 9: Mean weighted-mutual-overlap (w-m-o) scores for different values of $k$ on two corpora.

The general co-ranking model CorGen, by incorporating the conversation structure, outperforms the baselines in both metrics for all $k$ on blog ($p<0.05$), but fails to do so in many cases on email. On blog, there is also no significant difference between BiasRW and CorGen in w-m-o for all $k$ (Table 9), but CorGen outperforms BiasRW in w-s-m-o (Table 10) for higher values of $k$ (2,3,4,5) ($p<0.05$). On the other hand, on email, BiasRW always outperforms CorGen in both metrics for all $k$ ($p<0.05$). So we can conclude that, on blog, exploiting the conversation structure seems to be more beneficial than the leading sentences, whereas on email, we observe the opposite. The reason could be that the topic segments in blog are much longer than those of email (average length 27.16 vs. 12.6). Therefore, the FQGs of blog segments are generally larger and capture more information than the FQGs of email segments. Besides, email discussions are more focused than blog discussions. The leading sentences in email segments carry more informative clues than that of blog segments. This is also confirmed in Figure 9, where the leading sentences in email cover more of the human-authored words than they do in blog.

| | | k=1 | | k=2 | | k=3 | | k=4 | | k=5 | |
|---|---|---|---|---|---|---|---|---|---|---|---|
| | | Email | Blog | Email | Blog | Email | Blog | Email | Blog | Email | Blog |
| **Baselines** | FreqBL | 23.36 | 23.52 | 20.50 | 21.03 | 19.82 | 20.18 | 18.47 | 19.58 | 17.81 | 19.27 |
| | Lead-BL | 24.99 | 21.19 | 21.69 | 20.61 | 20.40 | 19.49 | 19.57 | 18.98 | 19.17 | 18.71 |
| **Models** | M&T | 18.71 | 22.08 | 16.25 | 19.59 | 14.62 | 17.91 | 14.29 | 17.27 | 14.06 | 16.92 |
| | Maui | 14.79 | 14.14 | 13.76 | 13.67 | 13.03 | 12.87 | 12.69 | 12.10 | 11.73 | 11.52 |
| | BiasRW | **28.87** | 24.63 | 24.76 | 22.51 | 22.48 | 21.36 | 21.67 | 20.95 | 21.28 | 20.78 |
| | BiasRW+ | 27.96 | 24.51 | 24.71 | 23.05 | 22.56 | 22.88 | 21.19 | 22.08 | 20.82 | 21.73 |
| | CorGen | 23.66 | 24.69 | 21.97 | 23.83 | 21.51 | 22.86 | 20.98 | 22.37 | 20.44 | 22.22 |
| | CorGen+ | 23.50 | 24.30 | 22.09 | **24.35** | 21.96 | **23.89** | 21.36 | 23.42 | 20.90 | 23.00 |
| | CorBias | 28.44 | **25.66** | **26.39** | 24.15 | 24.47 | 23.18 | **23.70** | 22.76 | **23.56** | 22.67 |
| | CorBias+ | 27.97 | 25.26 | 26.34 | 24.19 | **24.69** | 23.60 | 23.65 | **23.44** | 23.23 | **23.20** |

Table 10: Mean weighted-semantic-mutual-overlap scores for different values of $k$ on two corpora.

By combining the two forms of conversation specific information into a single model, CorBias delivers improved performance over CorGen and BiasRW in both metrics. On email, CorBias is significantly better than CorGen for all $k$ in both metrics ($p<0.01$). On blog, CorBias gets significant improvement over BiasRW for higher values of $k$ ($3, 4, 5$) in both metrics ($p<0.05$). The two sources of information are complementary and help each other to overcome the domain-specific limitations of the respective models. Therefore, one should exploit both information sources to build a generic domain-independent system.

When we include the conversation-level phrases (+ versions), we get a significant improvement in w-m-o on blog ($p<0.01$), but not on email. This may be because blog conversations have many more topical segments than email conversations (average topic number 10.77 vs. 2.5). Thus, there is little information for the label of a topical segment outside that segment in email conversations. However, note that including conversation-level phrases does not hurt the performance significantly in any case.

To further analyze the performance, Table 11 shows the mean w-m-o scores when only the best of $k$ output labels is considered. This allows us to judge the models' ability to generate the best label in the top $k$ list. The results are much clearer here. Generally speaking, among the models that do not include conversation-level phrases, CorBias is the best model, while including conversation-level phrases improves the performance further.

Table 12 shows some of the examples from our test set where the system-generated (i.e., CorBias+) labels are very similar to the human-authored ones. There are also many cases like the ones in Table 13, where the system-generated labels are reasonable, although they get low w-m-o and w-s-m-o scores when compared with the human-authored labels.

| | | k=2 | | k=3 | | k=4 | | k=5 | |
|---|---|---|---|---|---|---|---|---|---|
| | | Email | Blog | Email | Blog | Email | Blog | Email | Blog |
| **Baselines** | FreqBL | 27.02 | 23.69 | 29.79 | 24.29 | 31.12 | 24.88 | 31.25 | 25.58 |
| | LeadBL | 28.72 | 21.69 | 30.86 | 23.14 | 31.99 | 24.19 | 31.99 | 25.33 |
| **Models** | M&T | 21.45 | 21.70 | 23.12 | 23.18 | 25.23 | 23.82 | 25.45 | 24.07 |
| | Maui | 14.00 | 14.85 | 15.57 | 17.33 | 17.15 | 19.23 | 18.40 | 20.03 |
| | BiasRW | 29.34 | 24.92 | 31.42 | 25.18 | 32.58 | 25.89 | 32.97 | 26.64 |
| | BiasRW+ | 29.47 | 25.88 | 31.43 | 27.38 | 32.96 | 28.47 | 33.87 | 29.17 |
| | CorGen | 23.45 | 25.05 | 28.44 | 25.72 | 30.10 | 26.40 | 30.33 | 27.10 |
| | CorGen+ | 24.56 | 25.87 | 28.46 | 26.61 | 31.14 | 27.63 | 32.91 | 28.50 |
| | CorBias | 28.98 | 25.27 | 30.90 | 26.41 | 32.24 | 27.14 | 33.25 | 27.65 |
| | CorBias+ | **29.76** | **25.96** | **31.04** | **27.65** | **33.61** | **28.63** | **35.35** | **29.58** |

Table 11: Mean weighted-mutual-overlap (w-m-o) scores when the best of $k$ labels is considered.

| | Human-authored | System-generated (top 5) |
|---|---|---|
| **Email** | *Details of Bristol meeting* | *Bristol, face2face meeting, England, October* |
| | *Nashville conference* | *Nashville conference, Courseware developers, mid October, event* |
| | *Meeting agenda* | *detailed agenda, main point, meetings, revision, wcag meetings* |
| | *Design guidelines* | *general rule, design guidelines, accessible design, absolutes, forbid* |
| | *Contact with Steven* | *Steven Pemberton, contact, charter, status, schedule w3c* |
| **Blog** | *faster than light (FTL) travel* | *FTL travel, need FTL, limited FTL, FTL drives, long FTL* |
| | *Dr. Paul Laviolette* | *Dr. Paul Laviolette, bang theory, systems theory, extraterrestial beacons, laugh* |
| | *Vietnam and Iraq warfare* | *Vietnam war, incapable guerrilla war, war information, war ii, vietnamese war* |
| | *Pulsars* | *mean pulsars, pulsars slow time, long pulsars, relative pulsars, set pulsars* |
| | *Linux distributions* | *linux distro, linux support, major linux, viable linux* |

Table 12: Examples of Human-authored labels and System-generated labels.

| Human-authored | System-generated |
|---|---|
| *Meeting time and place* | *October, mid October, timing, w3c timing issues, Ottawa* |
| *Archaeology* | *religious site, burial site, ritual site, barrows tomb* |
| *Bio of Al* | *Al Gilman, standards reformer, repair interest group, ER IG, ER teams* |
| *Budget Constraints* | *budget, notice, costs, smaller companies, travel* |
| *Food choice* | *roast turkey breast, default choices, small number, vegetable rataouille, lunch* |

Table 13: Examples of System-generated labels that are reasonable but get low scores.

This is because most of the human-authored labels in our corpora are abstractive in nature. Annotators often write their own labels rather than simply copying keyphrases from the text. In doing so, they rely on their expertise and general world knowledge that may go beyond the contents of the conversation. In fact, although annotators reuse many words from the conversation, only 9.81% of the human-authored labels in blog and 12.74% of the human-authored labels in email appear verbatim in their respective conversations. Generating human-like labels will require a deeper understanding of the text and robust textual inference, for which our extractive approach can provide some useful input.

### 5.3 Full System Evaluation

In this section we present the performance of our end-to-end system. We first segment a given asynchronous conversation using our best topic segmenter (the supervised model), and then feed its output to our best topic labeler (the CorBias+ model). Table 14 presents the human agreement and the agreement of our system with the human annotators based on the best of $k$ outputs. For each system annotation we measure its agreement in w-m-o and w-s-m-o with the three human annotations using the method described in Section 4.3.3.

|  |  | System | | | | | Human |
|---|---|---|---|---|---|---|---|
|  |  | k=1 | k=2 | k=3 | k=4 | k=5 |  |
| **Email** | Mean w-m-o | 19.19 | 23.62 | 26.19 | 27.06 | 28.06 | 36.84 |
|  | Max w-m-o | 100.0 | 100.0 | 100.0 | 100.0 | 100.0 | 100.0 |
|  | Mean w-s-m-o | 24.98 | 32.08 | 34.63 | 36.92 | 38.95 | 42.54 |
|  | Max w-s-m-o | 108.43 | 108.43 | 108.43 | 108.43 | 108.43 | 107.31 |
| **Blog** | Mean w-m-o | 9.71 | 11.71 | 14.55 | 15.83 | 16.72 | 19.97 |
|  | Max w-m-o | 26.67 | 26.67 | 35.00 | 35.00 | 35.00 | 54.17 |
|  | Mean w-s-m-o | 15.46 | 19.77 | 23.35 | 25.57 | 26.23 | 28.22 |
|  | Max w-s-m-o | 47.10 | 47.28 | 47.28 | 48.54 | 48.54 | 60.76 |

Table 14: Performance of the end-to-end system and human agreement.

Notice that in email, our system gets 100% agreement in w-m-o metric for some conversations. However, there is a substantial gap between the mean and the max w-m-o scores. Similarly, in w-s-m-o, our system achieves a maximum of 108% agreement, but the mean varies from 25% to 39% depending on different values of $k$. In blog, the w-m-o and w-s-m-o scores are much lower. The maximum scores achieved in w-m-o and w-s-m-o metrics in blog are only 35% and 49% (for $k = 5$), respectively. The mean w-m-o score varies from 10% to 17%, and the mean w-s-m-o score varies from 15% to 28% for different values of $k$. This demonstrates the difficulties of topic segmentation and labeling tasks in blog conversations.

Comparing with Table 11, we can notice that inaccuracies in the topic segmenter affects the overall performance. However, our results are encouraging. Even though for lower values of $k$ there is a substantial gap between our results and the human agreement, as the value of $k$ increases, our results get closer to the human agreement, especially in w-s-m-o.

## 6. Conclusion and Future Direction

This work presents two new corpora of email and blog conversations annotated with topics, which, along with the proposed metrics, will allow researchers to evaluate their work quantitatively. We also present a complete computational framework for topic segmentation and labeling in asynchronous conversation.[24] Our approach extends state-of-the-art methods by considering the fine-grained structure of the asynchronous conversation, along with other conversational features. We do this by applying recent graph-based methods for NLP such as min-cut and random walk on paragraph, sentence or word graphs.

For topic segmentation, we extend the LDA and LCSeg unsupervised models to incorporate the fine-grained conversational structure (the Fragment Quotation Graph (FQG)), generating two novel unsupervised models LDA+FQG and LCSeg+FQG. In addition to that, we also present a novel graph-theoretic supervised segmentation model that combines lexical, conversational and topic features. For topic labeling, we propose two novel random walk models that extract the most representative keyphrases from the text, by respectively capturing conversation specific clues from two different sources: the leading sentences and the fine conversational structure (i.e., the FQG).

Experimental results in the topic segmentation task demonstrate that both LDA and LCSeg benefit significantly when they are extended to consider the FQG, with LCSeg+FQG being the best unsupervised model. The comparison of the supervised segmentation model with the unsupervised models shows that the supervised method outperforms the unsupervised ones even using only a few labeled conversations, being the best segmentation model overall. The outputs of LCSeg+FQG and the supervised model are also highly correlated with human annotations in both local and global metrics. The experiment on the topic labeling task reveals that the random walk models perform better when they exploit the conversation specific clues and the best results are achieved when all the sources of clues are exploited. The evaluation of the complete end-to-end system also shows promising results when compared with human performance.

This work can be extended in many ways. Given that most of the human-authored labels are abstractive in nature, we plan to extend our labeling framework to generate more abstract human-like labels that could better synthesize the information expressed in a topic segment. A promising approach would be to rely on more sophisticated methods for information extraction, combined with more semantics (e.g., phrase entailment) and data-to-text generation techniques. Another interesting venue for future work is to perform a more extrinsic evaluation of our methods. Instead of testing them with respect to a human gold standard, it would be extremely interesting to see how effective they are when used to support other NLP tasks, such as summarization and conversation visualization. We are also interested in the future to transfer our approach to other similar domains by domain adaptation methods. We plan to work on both synchronous and asynchronous domains.

---

24. Our annotated corpora, annotation manual and source code will be made publicly available from www.cs.ubc.ca/labs/lci/bc3.html

## Bibliographic Note

Portions of this work were previously published in two conference proceedings (Joty et al., 2010, 2011). This article significantly extends our previous work in several ways, most notably: (i) we complete the topic modeling pipeline by presenting a novel topic labeling framework (Section 3.2), (ii) we propose a new set of metrics for the topic labeling task (Section 5.2), (iii) we present a new annotated corpus of blog conversations, and show how the topic segmentation and labeling models perform on this new dataset (Section 4 and 5), and (iv) we demonstrate the performance of the end-to-end system (Section 5.3).

## Acknowledgments

This work was conducted at the University of British Columbia. We acknowledge the funding support of NSERC Canada Graduate Scholarship (CGS-D), NSERC BIN Strategic Network and NSERC discovery grant. We are grateful to the annotators for their great effort. Many thanks to Gabriel Murray, Jackie Cheung, Yashar Mehdad, Shima Gerani, Kelsey Allen and the anonymous reviewers for their thoughtful suggestions and comments.

## Appendix A. Metrics for Topic Segmentation

### A.1 One-to-One Metric

Consider the two different annotations of the same conversation having 10 sentences (denoted by colored boxes) in Figure 11(a). In each annotation, the topics are distinguished by different colors. For example, the *model output* has four topics, whereas the *human annotation* has three topics. To compute one-to-one accuracy, we take the model output and map its segments optimally (by computing the optimal max-weight bipartite matching) to the segments of the gold-standard human annotation. For example, the red segment in the model output is mapped to the green segment in the human annotation. We transform the model output based on this mapping and compute the percentage of overlap as the one-to-one accuracy. In our example, seven out of ten sentences overlap, therefore, the one-to-one accuracy is 70%.

### A.2 $Loc_k$ Metric

Consider the model output (at the left most column) and the human annotation (at the right most column) of the same conversation having 5 sentences (denoted by colored boxes) in Figure 12. Similar to Figure 11, the topics in an annotation are distinguished using different colors. Suppose we want to measure the $loc_3$ score for the fifth sentence (marked with yellow arrows at the bottom of the two annotations). In each annotation, we look at the previous 3 sentences and transform them based on whether they have *same* or *different* topics. For example, in the model output one of the previous three sentences is *same* (red), and in the human annotation two of the previous three sentences are *same* (green), when compared with the sentence under consideration. In the transformed annotations, *same* topics are denoted by gray boxes and *different* topics are denoted by black boxes. We compute $loc_3$ by measuring the overlap of the *same* or *different* judgments in the 3-sentence window. In our example, two of three overlap, therefore, the $loc_3$ agreement is 66.6%.

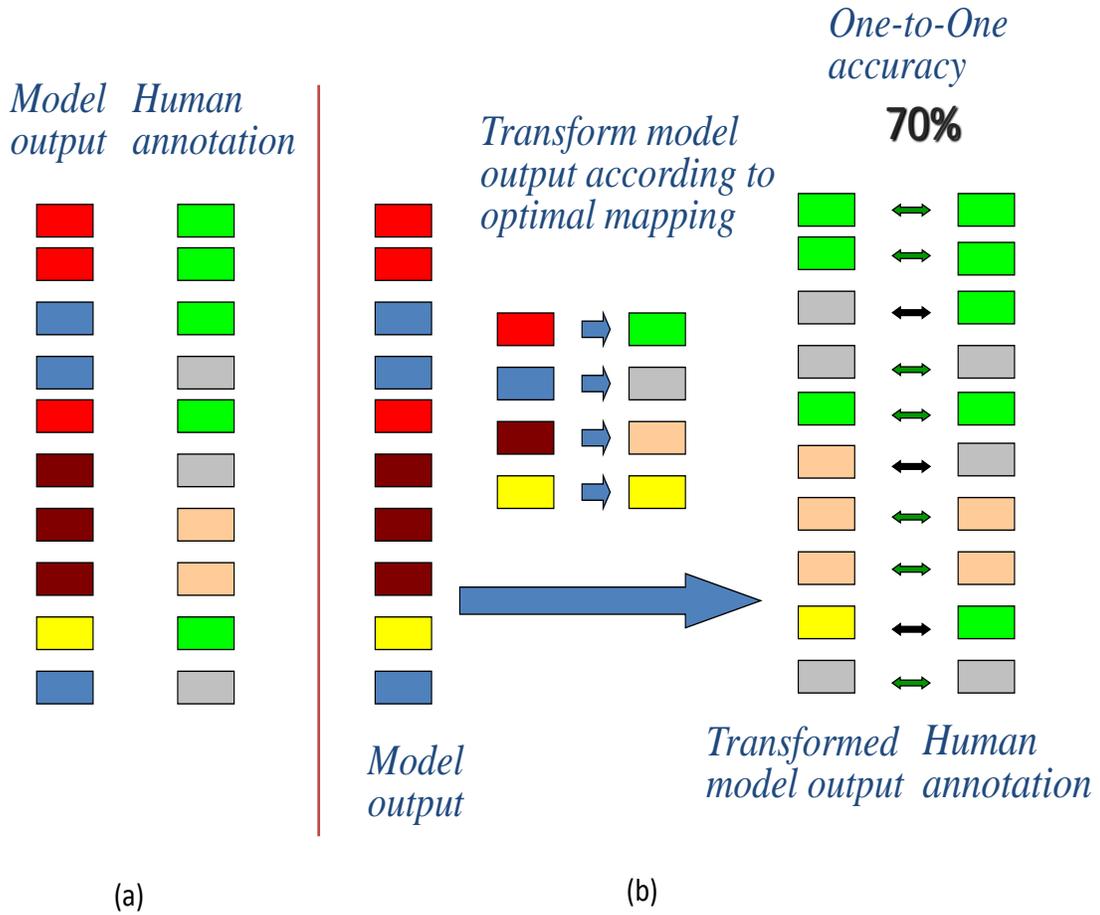

Figure 11: Computing one-to-one accuracy.

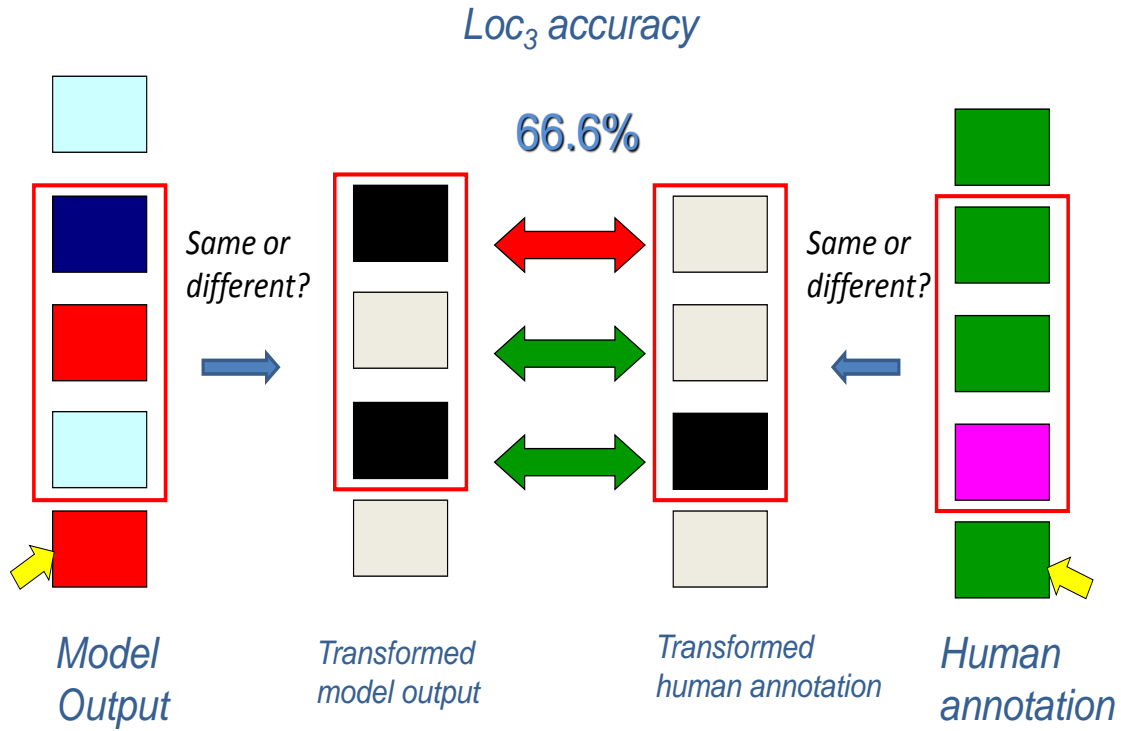

Figure 12: Computing $loc_3$ accuracy.